\documentclass{article}

\usepackage{microtype}
\usepackage{graphicx}
\usepackage{subfigure}
\usepackage{booktabs} 
\usepackage{multirow}
\usepackage{hyperref}

\usepackage[preprint]{icml2026}

\usepackage{amsmath}
\usepackage{amssymb}
\usepackage{mathtools}
\usepackage{amsthm}

\usepackage[capitalize,noabbrev]{cleveref}

\theoremstyle{plain}

\theoremstyle{definition}

\theoremstyle{remark}

\usepackage[textsize=tiny]{todonotes}

\icmltitlerunning{Sequence Diffusion Model for Temporal Link Prediction in Continuous-Time Dynamic Graph}

\begin{document}

\twocolumn[
\icmltitle{Sequence Diffusion Model for Temporal Link Prediction in Continuous-Time Dynamic Graph}




\begin{icmlauthorlist}
\icmlauthor{Nguyen Minh Duc}{yyy}
\icmlauthor{Viet Cuong Ta}{yyy}
\end{icmlauthorlist}

\icmlaffiliation{yyy}{VNU University of Engineering and Technology, Hanoi}

\icmlcorrespondingauthor{Viet Cuong Ta}{cuongtv@vnu.edu.vn}

\icmlkeywords{Dynamic Graphs, Link Prediction, Diffusion Model}

\vskip 0.3in
]



\printAffiliationsAndNotice{}  

\begin{abstract}
Temporal link prediction in dynamic graphs is a fundamental problem in many real-world systems. Existing temporal graph neural networks mainly focus on learning representations of historical interactions. Despite their strong performance, these models are still purely discriminative, producing point estimates for future links and lacking an explicit mechanism to capture the uncertainty and sequential structure of future temporal interactions.
In this paper, we propose \textbf{SDG}, a novel sequence-level diffusion framework that unifies dynamic graph learning with generative denoising. Specifically, SDG injects noise into the entire historical interaction sequence and jointly reconstructs all interaction embeddings through a conditional denoising process, thereby enabling the model to capture more comprehensive interaction distributions. To align the generative process with temporal link prediction, we employ a cross-attention denoising decoder to guide the reconstruction of the destination sequence and optimize the model in an end-to-end manner. Extensive experiments on various temporal graph benchmarks show that SDG consistently achieves state-of-the-art performance in the temporal link prediction task.
\end{abstract}

\section{Introduction}
Many real-world systems, such as social networks \cite{social}, recommendation systems \cite{dgs}, and citation networks \cite{citation}, can be naturally represented as dynamic graphs, where nodes denote entities and edges represent interactions, each annotated with a timestamp \cite{survey3}. A fundamental challenge in dynamic graphs is temporal link prediction, where the algorithm has to predict future connections of nodes based on their historical records \cite{tlp}, which underpins numerous applications, including recommendation, anomaly detection \cite{slade}, and knowledge graph completion \cite{tipnn}.

Existing temporal graph neural networks (TGNNs) mainly focus on representation learning for dynamic graphs. Memory-based models such as JODIE \cite{jodie}, DyRep \cite{dyrep}, and TGN \cite{tgn} model temporal evolution through recurrent updates of node embeddings, while memory-free methods such as TGAT \cite{tgat}, DyGFormer \cite{dygformer}, and CRAFT \cite{craft} directly aggregate temporal neighbors through attention or cross-attention mechanisms to produce the node representation. Despite their success, these approaches remain fundamentally discriminative \cite{dg-gen} and exhibit two key limitations.
First, these methods lack an explicit mechanism to represent uncertainty in the highly stochastic and noisy evolution of temporal interactions \cite{tspear}. Second, existing TGNNs optimize a point-wise loss function over individual future links, rather than modeling the sequential structure of future interaction. This limits their ability to reason about the temporal structure across multiple interactions.

To address these limitations, we draw inspiration from diffusion models \cite{ddpm,sde}, which provide probabilistic frameworks that naturally capture uncertainty and produce diversified generations through explicit learning of latent data distributions. This makes diffusion models emerge as powerful generative methods in many domains, such as computer vision \cite{ddpm,sde}, natural language processing \cite{diffusionlm, diffuseq}, and time-series modeling \cite{timedart, timediff}. However, it remains an open question on how to build a robust diffusion model for modeling dynamic graphs.

In this paper, we propose \textbf{S}equence \textbf{D}iffusion for Dynamic \textbf{G}raphs (SDG), a novel framework that unifies temporal graph learning with diffusion process. SDG injects noise into both historical neighbor interactions and the future destination, turning temporal link prediction into a sequence-level denoising task in which every interaction in the sequence is reconstructed from its noisy version. This encourages the model to learn richer node distributions rather than focusing only on the final interactions. SDG employs a scalable cross-attention denoise decoder, which leverages preceding sequence information to guide the denoising process. We summarize our main contributions as follows:
\begin{itemize}
\item We introduce a diffusion framework for continuous-time dynamic graphs. By injecting noise into the historical neighbors of the source node and the destination node, SDG can effectively learn richer node distributions.
\item We employ a cross-attention denoising decoder that conditions on the encoded temporal interactions to reconstruct the destination sequence, aligning the generative denoising process with temporal link prediction.
\item We conduct extensive experiments on multiple benchmark datasets, demonstrating that SDG consistently outperforms state-of-the-art temporal graph methods in temporal link prediction tasks.
\end{itemize}

\section{Related Works}

\subsection{Continuous-time Dynamic Graph Learning}
Dynamic graph learning aims to model the evolving patterns of node interactions over time, with temporal link prediction being the core task \cite{edgebank}. Early methods such as JODIE \cite{jodie}, DyRep \cite{dyrep}, and TGN \cite{tgn} adopt a memory-based architecture, where RNN models \cite{rnn} update node states as new edges arrive. Subsequent work moves beyond memory mechanisms, learning time-dependent node representations directly through temporal neighbor aggregation. These models include attention-based approaches such as TGAT \cite{tgat}, DyGFormer \cite{dygformer}, and TCL \cite{tcl}; as well as MLP-based designs such as GraphMixer \cite{graphmixer}, which leverages the MLP-Mixer \cite{mlpmixer} to capture temporal dependencies efficiently. Recently, CRAFT \cite{craft} introduced a cross-attention mechanism that enables target-aware modeling between candidate destinations and the source’s interaction history, addressing the challenge of unseen edge prediction. Although these advances have significantly improved temporal graph representation learning, they remain largely discriminative in nature. In contrast, our work explores diffusion-based generative modeling to capture uncertainty and long-range temporal dependencies, offering a new perspective for dynamic graph learning.

\subsection{Diffusion Models}
Diffusion models have recently emerged as a powerful class of generative models, achieving state-of-the-art results in domains such as computer vision \cite{ddpm, sde} and natural language processing \cite{diffusionlm}. The core idea is to learn a denoising process that reverses a forward noise injection, enabling the generation of complex distributions with high fidelity and flexibility. Building on this success, several works have explored diffusion models in the context of graphs. For example, GraphGDP \cite{gdb} and GraphDiffusion \cite{graphdiff} generate molecular and static graph structures by applying diffusion processes directly on adjacency matrices or node embeddings. These approaches primarily focus on static graphs, emphasizing structural generation rather than temporal dynamics.
Diffusion has also been applied to sequential data such as text and time-series. For instance, Diffusion-LM \cite{diffusionlm} and DiffuSeq \cite{diffuseq} adapt the denoising paradigm to language modeling and sequence generation, while DiffuRec \cite{diffurec} employs diffusion for recommendation sequences. In the time-series domain, methods such as TimeDiff \cite{timediff} and TimeDart \cite{timedart} use diffusion-style denoising for forecasting and representation learning. Despite these advances, the integration of diffusion models into dynamic graph learning remains largely unexplored. To our knowledge, the only work that leverages diffusion in this setting is CONDA \cite{conda}, which uses a diffusion process to generate augmented temporal views, rather than as a core predictive model. In contrast, our work proposes a sequence diffusion framework that explicitly models temporal neighbor interactions and uses the denoising process itself for temporal link prediction.

\section{Preliminaries}

\subsection{Problem Formulation}

\textbf{Continuous-time Dynamic Graph.} A continuous-time dynamic graph can be represented as a chronological sequence of interactions between specific pairs of nodes: $G=\{(u_1, v_1, t_1), \dots, (u_n, v_n, t_n)\}$, with $0 \le t_0 \le t_1 \le \dots \le t_n$, where $u_i, v_i \in \mathcal{V}$ denote the source node and destination node of the link $i^{th}$ at the timestamp $t_i \in \mathcal{T}$. $\mathcal{V}$ and $\mathcal{T}$ are the node set and timestamp set, respectively.

\textbf{Historical Interaction Sequence.} For a source node $u$ and time $t$, we define the notation $S_{u, t} = \{(v_1, t_1),  (v_2, t_2), \dots, (v_L, t_L)\}$ to denote the ordered sequence of neighbors that node $u$ interacted before timestamp $t$, with $t_1 \le t_2 \le \dots \le t_{L}  < t $ and $L$ is the maximum length of the sequence.

\textbf{Temporal Link Prediction.} Given a source node $u$, a candidate destination node $v$, a timestamp $t$, and the historical interaction sequences $S_{u, t}$. The temporal link prediction task is to estimate the likelihood of an interaction between $u$ and $v$ at time $t$. We follow the protocol in \cite{tgb-seq} to frame this task as a ranking problem, as real-world applications often require identifying the most likely destination from a large pool of candidates.

\textbf{Temporal Link Prediction as Conditional Generation.} We define the source node representation as $\mathbf{H}(u, t) \in \mathbb{R}^d$ and the destination node $\mathbf{H}(v, t) \in \mathbb{R}^d$, The temporal link prediction task can be reformulated as a conditional generation problem, where the destination representation is generated conditioned on historical interactions.
Specifically, our objective is to learn a conditional distribution $p_\theta(\mathbf{H}(v, t) \mid \mathbf{H}(S_{u,t}))$, where $\mathbf{H}(S_{u,t}) \in \mathbb{R}^{L \times d}$ is the sequence embedding for the interaction history of the source node up to time $t$.

\subsection{Conditional Diffusion Model}
\label{sec:diff_preliminary}
Based on the above formulation, we adopt the denoising diffusion probabilistic model (DDPM) \cite{ddpm} to parameterize the conditional distribution. We first denote the parameterization as $p_{\theta}(x^0)$, where $x^0$ is the target sample. In the context of temporal link prediction, we want to model the embedding of the destination $v$ by setting $x^0=\mathbf{H}(v, t)$.

To learn the $p_{\theta}(x^0)$, DDPM first gradually add Gaussian noise to $x^0$ as ${x^1, x^2, \dots x^K}$, where after $K$ steps, the final variable $x^K$ follows a Gaussian distribution $\mathcal{N}(0, I)$. Formally, this process is defined as a Markov chain:
\begin{equation}
q(x^k \mid x^{k-1}) = \mathcal{N}(x^k; \sqrt{1 - \beta_k} x^{k-1}, \beta_k I \bigr),
\label{eq:forwarddiff}
\end{equation}
where $\beta_k$ is the variance of the noise added at each step $k$ controlled by a schedule ${\beta_k \in (0, 1)}^{K}{k=1}$.
By defining the cumulative product of the noise scales as $\bar{\alpha}^k = \prod{k’=1}^{k} (1 - \beta^{k’})$, we can derive a closed-form expression for the marginal distribution of $x^k$ conditioned on the original clean data $x^0$ as:
\begin{align}
&q(x^k \mid x^0) = \mathcal{N}(x^k; \sqrt{\bar{\alpha}^k}x^0, (1 - \bar{\alpha}^k) I), \\
&x^k = \sqrt{\bar{\alpha}^k}x^0 + \sqrt{1 - \bar{\alpha}^k}\epsilon,
\quad \epsilon \sim \mathcal{N}(0, I),
\end{align}
In the reverse process, the goal is to train a parameterized denoise model that can invert this noising procedure conditioned on context $c$. In the context of dynamic graph learning, we rely on the historical interaction, more specifically $c = \mathbf{H}(S_{u, t})$
Each denoising step is modeled as a conditional Gaussian transition. Each denoising step can also be formulated as a Gaussian transition as:
\begin{align}
p_{\theta}(x^{k-1} | x^k, c) = \mathcal{N}(x^{k-1}; \mu_{\theta}(x^k, k, c), \sigma_k I),
\end{align}
where $\mu_{\theta}(x^k, k, c)$ is the predicted mean and $\sigma_k$ denotes a fixed variance term depending on step $k$. We can derive the mean $\mu_{\theta}$ in terms of the original data $x^0$:
\begin{align}
\mu_{\theta}(x^k, k, c) = \frac{\sqrt{1 - \beta^k}(1 - \bar{\alpha}^k)}{1 - \bar{\alpha}^k} x^k + \frac{\alpha^{k - 1}\beta^k}{1 - \bar{\alpha}^k} \hat{x}^0.
\label{eq:x0_mean}
\end{align}
Here, $\hat{x}^0 = f_{\theta}(x^k, k, c)$ is the output of the denoise model. The denoise model is usually trained by minimizing a mean-squared error (MSE) loss between the predicted clean sample $\hat{x}^0$ and the ground truth $x^0$, instead of the original Variational Lower Bound (VLB) as:
\begin{align}
\mathcal{L}(\theta) = \mathbb{E}_{k, x^0}\left[||x^0 - \hat{x}^0||^2_{2}\right].
\end{align}

\section{Methodology}
\begin{figure*}[h]
\centering
\includegraphics[width=1.0\linewidth]{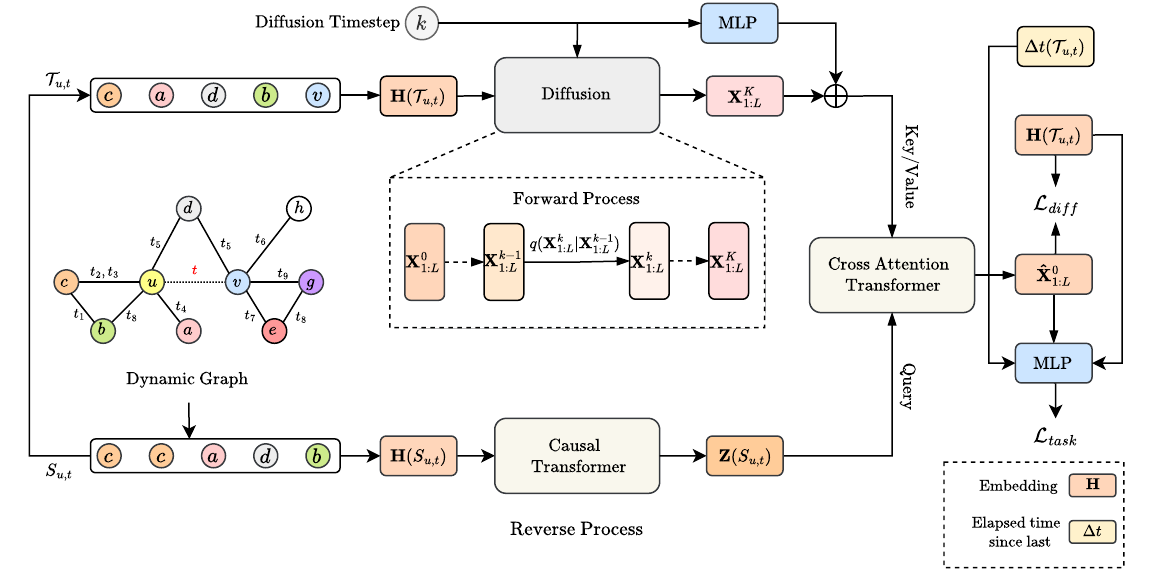}
\caption{The overall architecture of SDG, which injects noise into both historical neighbors and the future destination. SDG reconstructs the destination sequence through a cross-attention denoising decoder conditioning on the encoded history.}
\label{fig:pipeline}
\end{figure*}

We now present our SDG that extends the diffusion process for temporal link prediction (Figure \ref{fig:pipeline}). Given the historical sequence $S_{u,t}$, SDG first encodes the sequence into the latent space $\mathbf{Z}(S_{u,t})$, which serves as contextual guidance to reconstruct the target sequence $\mathcal{T}_{u,t}$. During inference, SDG iteratively generates the destination sequence from pure Gaussian noise, guided by the context embedding $\mathbf{Z}(S_{u,t})$ of the source node, and uses the last element of the generated sequence as the target node. By unifying historical context encoding and diffusion-based denoising, SDG captures long-range temporal dependencies and uncertainty in dynamic graphs within a single framework.

\subsection{Encoding Sequence Interactions}
Given an interaction $(u, v, t)$, we first extract the historical neighbors $S_{u, t}$ for source node $u$ at timestamp $t$. In practice, we keep the $L$ most recent 1-hop neighbors, which is sufficient for modeling temporal patterns \cite{dygformer}.

Since most temporal graph benchmarks lack explicit node features for the latent diffusion module to operate on, we assign the interaction sequence with a learnable node embedding lookup table $\mathbf{H} \in \mathbb{R}^{N \times d}$. The historical embedding of node $u$ is indicated as $\mathbf{H}(S_{u,t}) = [\mathbf{H}(v_1), \ldots, \mathbf{H}(v_L)]$, where we omit timestamps for simplicity as they are not used in the embedding lookup. We then add positional encoding to obtain the initial representation:
\begin{align}
\mathbf{Z}^{0}_{1:L}(S_{u, t})=\left[\mathbf{H}(v_1), \mathbf{H}(v_2), \cdots, \mathbf{H}(v_L)\right] + \mathbf{PE},
\end{align}
where $\mathbf{PE}$ denotes the sinusoidal positional encoding \cite{attention}. We can also obtain the destination embedding $\mathbf{H}(v)$ for prediction.

Next, we employ a causal Transformer model \cite{attention} to capture the evolving trends from left to right of the historical sequence $\mathbf{H}(S_{u, t}) \in \mathbb{R}^{L \times d}$. Each layer in the model applies multi-head self-attention with a causal mask, followed by a feed-forward network (FFN):
\begin{equation}
\mathbf{Z}_{1:L}(S_{u, t})
= \text{Transformer}(\mathbf{H}_{1:L}(S_{u, t}); \mathbf{M}),
\end{equation}
where $\mathbf{M}$ is the attention mask ensuring position $i$ only allows to attends to positions $\le i$. After $N_\text{layers}$ of Transformer blocks, the output $\mathbf{Z}_{1:L}(S_{u, t})$ of this module will be utilized as the condition signal guidance for the diffusion module during the denoising process.

\subsection{Sequence Diffusion Model}

Instead of adding noise solely to the final destination embedding $\mathbf{H}(v)$, we add noise to the historical sequence and the final destination embedding. Firstly, we shift the historical interactions by one step and append the destination node to form the target sequence $\mathcal{T}_{u,t} = \{(v_2, t_2), \dots, (v, t)\}$. Then, the embedding of this sequence can be represented as $\mathbf{H}(\mathcal{T}_{u, t}) = \left[\mathbf{H}_{2:L}(S_{u, t}), \mathbf{H}(v)\right] \in \mathbb{R}^{L \times d}$.

We set this sequence embedding as the clean embedding for our diffusion as $\mathbf{X}^0 = \mathbf{H}(\mathcal{T}_{u, t})$. Given $\mathbf{X}^0_{1:L}$, we instantiate the forward diffusion process from Section \ref{sec:diff_preliminary} at the sequence level. The noisy destination sequence $\mathbf{X}^k_{1:L}$ at timestep $k$ is given by:
\begin{equation}
\mathbf{X}^k_{1:L} = 
\sqrt{\bar{\alpha}^k}\,\mathbf{X}_{1:L}^0
+ \sqrt{1 - \bar{\alpha}^k}\,\boldsymbol{\epsilon},
\end{equation}
where $\boldsymbol{\epsilon}$ is sampled from $\mathcal{N}(\mathbf{0}, \mathbf{I})$ with the same shape as $\mathbf{X}^0_{1:L}$. Noising the entire sequence turns the diffusion objective into a sequence-level denoising problem where every position in $\mathcal{T}_{u,t}$ must be reconstructed from its noisy version, conditioning on the previous interactions instead of regressing a single vector $\mathbf{H}(v)$. This provides a richer supervision signal and effectively pushes the model to approximate a joint distribution over destination sequences.

The reverse process is now a conditional Markov chain over sequences, which can be defined as follows:
\begin{equation}
\begin{aligned}
p_\theta(\mathbf{X}_{1:L}^{k-1} \mid \mathbf{X}_{1:L}^k, \mathbf{Z}_{1:L}, k)
= \mathcal{N}\Big(
    \mathbf{X}_{1:L}^{k-1};\,
    \boldsymbol{\mu}_\theta(\mathbf{X}_{1:L}^k, \\\mathbf{Z}_{1:L}, k),\,
    \sigma_k^2 \mathbf{I}
\Big),
\end{aligned}
\end{equation}
where the mean follows the $x_0$-prediction from Equation \ref{eq:x0_mean}. In the next section, we provide details on the architecture of the sequence-level denoising network $f_{\theta}$:
\begin{equation}
\hat{\mathbf{X}}^0_{1:L}
= f_{\theta}(\mathbf{X}^k_{1:L}, \mathbf{Z}_{1:L}, k),
\end{equation}

\subsection{Denoising Network}
The denoising network $f_\theta$ jointly reconstructs the target sequence by taking the noisy target sequence $\mathbf{X}_{1:L}^k$, the diffusion step $k$ as input, and the historical context $\mathbf{Z}_{1:L}(S_{u,t})$ as conditional guidance. The timestep $k$ is first encoded using a sinusoidal time embedding $\gamma(\cdot)$ \cite{diffurec}, followed by an MLP. The resulting time embedding is added to each position of the noisy sequence:
\begin{equation}
    \hat{\mathbf{X}}^{k}_{1:L} = \mathbf{X}^{k}_{1:L} + \mathrm{MLP}\big(\gamma(k)\big).
\end{equation}
To model interaction evolution during denoising, we encode the trajectory context with a causal Transformer and use it as queries in cross-attention, while the time-conditioned noisy sequence serves as keys and values to refine noisy destinations. The forward block is presented below:
\begin{align}
\mathbf{Z}_{\text{ctx}} &= 
\mathrm{Transformer}\big(\mathbf{Z}_{1:L}(S_{u,t}), \mathbf{M}\big), \\
\hat{\mathbf{X}}_{1:L}^0 &= \mathrm{CrossTransformer}\big(\mathbf{Z}_{\text{ctx}},\hat{\mathbf{X}}_{1:L}^{k} \big),
\end{align}
where $\hat{\mathbf{X}}^0 \in \mathbb{R}^{L \times d}$ is the 
prediction by the network of the clean target sequence $\mathbf{X}^0$. We also set the number of blocks to $N_\text{layers}$. This design lets the history representation guide the denoising at each position, yielding destination embeddings that are temporally coherent and aligned with the observed interaction patterns.

\subsection{Training Objective}

To train the diffusion model, we adopt the $x_0$ prediction scheme rather than standard $\epsilon$ pipeline because the historical interactions contain highly irregular noisy components, thus estimating the noise $\epsilon$ can be more difficult.

Typically, diffusion models are trained using a mean squared error (MSE). However, \cite{graphmae, preferdiff} have shown that MSE is sensitive to vector norms and dimensionality, which is not optimal for our ranking phase. Therefore, we replace the standard MSE term in the ELBO with a cosine-based reconstruction error, which is invariant to embedding scale and better aligned with ranking objectives. We define the diffusion reconstruction loss as:
\begin{equation}
\mathcal{L}_{\text{diff}} = \frac{1}{L}\sum_{i=1}^L
    \bigl(1 - \cos(\hat{\mathbf{X}}^0_i,\mathbf{X}^0_i)\bigr)^2,
\label{eq:diff_loss}
\end{equation}
where $\cos(\cdot,\cdot)$ denotes cosine similarity and $\hat{\mathbf{X}}_i$ is the $i$-th element of $\hat{\mathbf{X}}^0$. We show in the Appendix \ref{sec:elbo} that this objective corresponds to a valid ELBO variant under a cosine-based similarity measure. This reconstruction loss encourages the diffusion model to match the ground-truth embeddings; however, it is not directly aligned with the ranking objective in temporal link prediction on its own.

To learn a predictive scoring function, we take the element-wise product between the denoised representation $\mathbf{X}^0$ and the destination node embedding, concatenated with temporal features. The final representation is fed to MLP to predict the edge likelihoods between the source and the destination.
\begin{align}
\hat{\mathbf{y}}_t = \text{MLP}\Big(\text{concat}(\mathbf{X}^0 \cdot \mathbf{H}(\mathcal{T}_{u, t}), \gamma(\Delta t(\mathcal{T}_{u, t})))\Big),
\label{eq:score}
\end{align}
where $\cdot$ denote the dot-product operation, $\hat{\mathbf{y}}_t \in \mathbb{R}^{L \times N}$ is the score matrix over sequence positions and candidate nodes, and the $i$-th row $\hat{\mathbf{y}}_{t, i} \in \mathbb{R}^{\mathcal{C}_{u,t}}$ contains the scores for all $\mathcal{C}_{u,t}$ candidate nodes at position $i$. During training, we adopt random negative sampling, where the candidate set sequence contains one positive sequence $\mathcal{T}^{+}_{u, t}$ and one negative sequence $\mathcal{T}^{-}_{u, t}$.

The term $\Delta t(\mathcal{T}_{u, t}) = t - t(\mathcal{T}_{u, t})$ is the elapsed time sequence between the prediction time $t$ and the last activity time of the destination sequence $\mathcal{T}_{u, t}$. $\gamma$ denotes a linear layer that projects $\Delta t$ to a time-context vector $\mathbb{R}^d$ to capture the temporal status of the destination, following \cite{craft}.

The scoring network is optimized with either binary cross-entropy (BCE) loss or Bayesian Personalized Ranking \cite{bpr} (BPR) loss. In the main text, we focus on the BCE case and defer the BPR formulation to the Appendix. To better exploit the sequential structure, we decompose the BCE loss into an intermediate-position loss $\mathcal{L}_{\text{inter}}$ and a final-position loss $\mathcal{L}_{\text{last}}$, yielding the overall task loss: 
\begin{align}
\mathcal{L_{\text{task}}}
= &\underbrace{
    - \log \sigma(\hat{y}^+_{t, L})
    - \log \big(1 - \sigma(\hat{y}^-_{t, L})\big)
}_{\mathcal{L}_{\text{last}}}
+ \\
&\lambda_{\text{inter}}\,
  \underbrace{
   \frac{1}{L - 1}\sum_{i=1}^{L-1}
      - \log \sigma(\hat{y}^+_{t, i})
      - \log \big(1 - \sigma(\hat{y}^-_{t, i})\big)
  }_{\mathcal{L}_{\text{inter}}},
\end{align}
where $\lambda_{\text{inter}}$ controls the relative contribution of the intermediate-position supervision. Finally, we combine the ranking loss with the diffusion reconstruction
loss to obtain the overall training objective:
\begin{equation}
\mathcal{L}
= \mathcal{L}_{\text{task}}
+ \lambda_{\text{diff}}\,\mathcal{L}_{\text{diff}},
\end{equation}
where $\lambda_{\text{diff}}$ balances the ranking objective and the diffusion reconstruction objective. In this way, the diffusion model is encouraged to produce denoised embeddings that are highly informative for ranking candidate nodes along the sequence. We train SDG in an end-to-end manner; the training procedure is shown in Algorithm \ref{alg:training}

\begin{table*}[h]
    \centering
    \caption{Results of SDG and baselines on seen-dominant dynamic graph datasets. The best and second-best results in each metric are marked in \textbf{bold} and \underline{underlined}, respectively.}
    \vspace{2mm}
    \label{tbl:seen}
    \resizebox{\linewidth}{!}{
    \begin{tabular}{lcccccccccc}
    \toprule
    \textbf{Datasets} & \multicolumn{2}{c}{\textbf{Wikipedia}} & \multicolumn{2}{c}{\textbf{Reddit}} & \multicolumn{2}{c}{\textbf{MOOC}} & \multicolumn{2}{c}{\textbf{Lastfm}} & \multicolumn{2}{c}{\textbf{UCI}} \\
    & MRR & HR@10 & MRR & HR@10 & MRR & HR@10 & MRR & HR@10 & MRR & HR@10 \\
    \midrule \midrule
    JODIE & 76.15{\scriptsize $\pm$0.57} & 84.45{\scriptsize $\pm$0.83} & 77.59{\scriptsize $\pm$0.70} & 87.74{\scriptsize $\pm$0.27} & 19.91{\scriptsize $\pm$3.06} & 29.87{\scriptsize $\pm$0.00} & 19.71{\scriptsize $\pm$0.62} & 28.79{\scriptsize $\pm$0.69} &  54.14{\scriptsize $\pm$2.18} & 70.81{\scriptsize $\pm$0.78} \\
    DyRep & 67.52{\scriptsize $\pm$1.10} & 78.20{\scriptsize $\pm$0.54} & 74.63{\scriptsize $\pm$1.02} & 85.94{\scriptsize $\pm$0.47}& 17.71{\scriptsize $\pm$1.27} & 35.09{\scriptsize $\pm$0.00}& 21.50{\scriptsize $\pm$1.51} & 28.29{\scriptsize $\pm$2.50}& 16.41{\scriptsize $\pm$0.70} & 24.38{\scriptsize $\pm$1.59} \\
    TGAT & 73.02{\scriptsize $\pm$0.64} & 79.98{\scriptsize $\pm$0.49} & 78.03{\scriptsize $\pm$0.25} & 86.55{\scriptsize $\pm$0.00} & 31.55{\scriptsize $\pm$4.14} & 41.02{\scriptsize $\pm$0.00} & 24.41{\scriptsize $\pm$0.86} & 29.44{\scriptsize $\pm$0.19} & 33.03{\scriptsize $\pm$0.38} & 34.50{\scriptsize $\pm$0.77} \\
    TGN & 81.22{\scriptsize $\pm$0.25} & 87.41{\scriptsize $\pm$0.13}& 79.25{\scriptsize $\pm$0.24} & 87.20{\scriptsize $\pm$0.00} &
    39.03{\scriptsize $\pm$4.53} & 41.78{\scriptsize $\pm$0.00} & 19.89{\scriptsize $\pm$4.20} & 27.19{\scriptsize $\pm$4.66} & 39.79{\scriptsize $\pm$8.14} & 49.15{\scriptsize $\pm$12.5} \\
    GraphMixer & 72.14{\scriptsize $\pm$0.68} & 79.84{\scriptsize $\pm$0.56} & 71.70{\scriptsize $\pm$0.06} & 84.03{\scriptsize $\pm$0.07} & 28.74{\scriptsize $\pm$0.27} & 37.07{\scriptsize $\pm$0.26} & 27.50{\scriptsize $\pm$0.44} & 35.60{\scriptsize $\pm$0.16}& 59.11{\scriptsize $\pm$1.45} & 72.90{\scriptsize $\pm$2.35} \\
    DyGFormer & \underline{88.81{\scriptsize $\pm$0.05}} & \underline{90.95{\scriptsize $\pm$0.10}} & 88.80{\scriptsize $\pm$0.01} & 94.09{\scriptsize $\pm$0.02} & 41.95{\scriptsize $\pm$0.01} & 52.85{\scriptsize $\pm$0.01} & 46.69{\scriptsize $\pm$0.19} & 62.35{\scriptsize $\pm$0.07} & \underline{75.73{\scriptsize $\pm$0.21}} & \textbf{82.39{\scriptsize $\pm$0.13}} \\
    
    CRAFT & {87.74}{\scriptsize $\pm$0.06} & 90.12{\scriptsize $\pm$0.34} & \underline{88.95{\scriptsize $\pm$0.04}} & \underline{94.39{\scriptsize $\pm$0.03}}& \underline{58.79{\scriptsize $\pm$0.03}} & \underline{78.68{\scriptsize $\pm$0.08}} & \underline{\textbf{54.53{\scriptsize $\pm$0.17}}} & \underline{\textbf{69.95{\scriptsize $\pm$0.02}}} & 72.44{\scriptsize $\pm$0.12} & 79.56{\scriptsize $\pm$0.20} \\
    SDG & \textbf{89.17{\scriptsize $\pm$0.06}} & \textbf{{91.40\scriptsize $\pm$0.10}} & \textbf{89.13{\scriptsize $\pm$0.08}} & \textbf{94.76{\scriptsize $\pm$0.10}} & \textbf{60.55{\scriptsize $\pm$0.08}} & \textbf{79.93{\scriptsize $\pm$0.02}} & \underline{53.79{\scriptsize $\pm$0.03}} & \underline{69.15{\scriptsize $\pm$0.05}} & \textbf{76.13{\scriptsize $\pm$0.15}} & \underline{79.78{\scriptsize $\pm$0.09}} \\
    \midrule
    Rel.Imprv. & 0.41\% & 0.71\% & 0.20\% & 0.39\% & 2.99\% & 1.59\% & -1.36\% & -1.14\% & 0.53\% & -3.16\% \\
    Abs.Imprv. & 0.37 & 0.63 & 0.18 & 0.37 & 1.76 & 1.25 & -0.74 & -0.80 & 0.40 & -2.61 \\
    \bottomrule
    \end{tabular}
    }
\end{table*}

\subsection{Inference Process}
During the inference stage, we first derive the historical representation of the source node $\mathbf{H}(S_{u, t})$. Starting from pure Gaussian noise, we then utilize the denoising network $f_\theta$ to iteratively generate latent embeddings, following the DDPM samplers \cite{ddpm}, until the inferred sequence embedding $\mathbf{\hat{X}}^0$ is obtained. More details can be found in Algorithm \ref{alg:inference}. Finally, we take out the last position of the destination embedding to produce the ranking scores for each destination node in the candidate set.

\section{Experiments}
In this section, we verify the effectiveness and efficiency of SDG across a variety of dynamic graph datasets on the temporal link prediction task. We provide a series of comprehensive and detailed analyses that demonstrate the superior experimental results of our proposed model.
Our code is available at this repository url\footnote{The code will be available upon acceptance}.

\begin{table*}[h]
    \centering
    \caption{Results of SDG and baselines on unseen-dominant datasets. The best and second-best results in each metric are marked in \textbf{bold} and \underline{underlined}, respectively. OOT denotes Out-Of-Time problem.}
    \vspace{2mm}
    \label{tbl:unseen}
    \resizebox{\linewidth}{!}{
    \begin{tabular}{lcccccccccc}
    \toprule
    \textbf{Datasets} & \multicolumn{2}{c}{\textbf{GoogleLocal}} & \multicolumn{2}{c}{\textbf{YouTube}} & \multicolumn{2}{c}{\textbf{Flickr}} & \multicolumn{2}{c}{\textbf{ML-20M}} & \multicolumn{2}{c}{\textbf{Taobao}} \\
    & MRR & HR@10 & MRR & HR@10 & MRR & HR@10 & MRR & HR@10 & MRR & HR@10 \\
    \midrule \midrule
    JODIE &
    42.81{\scriptsize $\pm$0.50} & 58.17{\scriptsize $\pm$0.35} & 45.44{\scriptsize $\pm$1.07} & 61.63{\scriptsize $\pm$1.31} & 43.84{\scriptsize $\pm$3.61} & 64.34{\scriptsize $\pm$7.40} & 20.30{\scriptsize $\pm$0.68} & 30.44{\scriptsize $\pm$1.35} & 46.99{\scriptsize $\pm$1.72} & 64.07{\scriptsize $\pm$1.65} \\
    DyRep &
    36.52{\scriptsize $\pm$0.12} & 50.48{\scriptsize $\pm$0.36} & 42.10{\scriptsize $\pm$5.42} & 58.64{\scriptsize $\pm$1.11} & 41.88{\scriptsize $\pm$4.67} & 66.15{\scriptsize $\pm$3.62} & 21.28{\scriptsize $\pm$0.29} & 32.39{\scriptsize $\pm$0.34} & 38.67{\scriptsize $\pm$1.35} & 53.57{\scriptsize $\pm$1.03}\\
    TGAT &
    16.65{\scriptsize $\pm$0.32} & 24.52{\scriptsize $\pm$0.53} & 40.31{\scriptsize $\pm$2.09} & 49.78{\scriptsize $\pm$2.25} & 22.81{\scriptsize $\pm$4.55} & 29.66{\scriptsize $\pm$5.47} & OOT & OOT & OOT & OOT \\
    TGN &
    53.97{\scriptsize $\pm$1.19} & 71.48{\scriptsize $\pm$1.32} & \underline{58.95{\scriptsize $\pm$5.07}} & \textbf{71.61{\scriptsize $\pm$3.46}} &
    52.74{\scriptsize $\pm$2.74} & 76.07{\scriptsize $\pm$0.86} & 24.39{\scriptsize $\pm$1.45} & 37.84{\scriptsize $\pm$2.06} & 60.18{\scriptsize $\pm$0.14} & 75.59{\scriptsize $\pm$0.08}\\
    GraphMixer &
    20.24{\scriptsize $\pm$0.22} & 28.16{\scriptsize $\pm$0.39} & 57.75{\scriptsize $\pm$0.11} & 68.45{\scriptsize $\pm$0.03} & 44.10{\scriptsize $\pm$0.06} & 64.15{\scriptsize $\pm$0.05} & 21.72{\scriptsize $\pm$0.65} & 33.38{\scriptsize $\pm$1.00} & 30.96{\scriptsize $\pm$0.13} & 43.37{\scriptsize $\pm$0.05}\\
    DyGFormer &
    15.76{\scriptsize $\pm$0.98} & 22.02{\scriptsize $\pm$1.46} & 47.04{\scriptsize $\pm$3.06} & 57.72{\scriptsize $\pm$3.56} & 43.40{\scriptsize $\pm$0.49} & 54.92{\scriptsize $\pm$1.30} & OOT & OOT & OOT & OOT \\
    CRAFT & 
    \underline{54.68{\scriptsize $\pm$0.58}} & \underline{72.24{\scriptsize $\pm$0.69}} & 55.62{\scriptsize $\pm$0.08} & 65.47{\scriptsize $\pm$0.04}& \underline{61.27{\scriptsize $\pm$0.28}} & \underline{79.11{\scriptsize $\pm$0.06}} & \underline{36.01{\scriptsize $\pm$0.19}} & \underline{52.28{\scriptsize $\pm$0.07}} & 
    \underline{67.41{\scriptsize $\pm$0.14}} & 
    \underline{79.08{\scriptsize $\pm$0.15}}\\
    SDG &
    \textbf{62.60{\scriptsize $\pm$0.62}} & \textbf{78.54{\scriptsize $\pm$0.43}} &
    \textbf{60.54{\scriptsize $\pm$0.36}} & \underline{71.01{\scriptsize $\pm$0.12}} & \textbf{61.79{\scriptsize $\pm$0.10}} & \textbf{80.92{\scriptsize $\pm$0.18}} & \textbf{36.63{\scriptsize $\pm$0.06}} & \textbf{53.55{\scriptsize $\pm$0.19}} & 
    \textbf{69.70{\scriptsize $\pm$0.19}} & 
    \textbf{81.43{\scriptsize $\pm$0.19}} \\
    \midrule
    Rel.Imprv. & 14.48\% & 8.72\% & 2.70\% & -0.84\% & 0.84\% & 2.29\% & 1.72\% & 2.43\% & 3.40\% & 2.97\%\\
    Abs.Imprv. & 7.92 & 6.30 & 1.59 & -0.60 & 0.52 & 1.81 & 0.62 & 1.27 & 2.29 & 2.35 \\
    \bottomrule
    \end{tabular}
    }
\end{table*}

\subsection{Experimental Setup}
\textbf{Datasets} We assess the ability of SDG in performing temporal link prediction with 10 datasets covering various domains and scales. We first test our method on five commonly used small datasets \cite{dygformer}: Wikipedia, Reddit, MOOC, Lastfm, and UCI. To further verify the scalability of SDG, we evaluate SDG on a large-scale benchmark TGB-Seq, including GoogleLocal, YouTube, Flickr, ML-20M, and Taobao. The details about the dataset statistics and descriptions are given in Appendix \ref{sec:data_detail}.

\textbf{Baselines} We compare our method to several state-of-the-art continuous-time temporal graph baselines, including JODIE \cite{jodie}, DyRep \cite{dyrep}, TGAT \cite{tgat}, TGN \cite{tgn}, GraphMixer \cite{graphmixer}, DyGFormer \cite{dygformer}, CRAFT \cite{craft}. The details about the baseline descriptions and their hyperparameters are given in Appendix \ref{sec:baseline_detail}.

\noindent
\textbf{Evaluation Protocol} For TGB-Seq benchmark datasets, we use their original split for training, validation, and testing, with their predefined negative samples for each set \cite{tgb-seq}.
For the remaining five common datasets, we chronologically split each dataset into training, validation, and test sets with 75\%, 15\%, and 15\% of the edges, respectively.
We evaluate the ranking performance using two metrics: Mean Reciprocal Rank (MRR) and Hit Rate (HR@$K$) with $K = 10$. HR@K reflects the model’s retrieval ability, while MRR indicates its ranking performance.

\textbf{Implementation Details} To ensure a fair comparison, we use TGB-Seq \cite{tgb-seq} and DyGLib \cite{dygformer} to reproduce all baselines via the same training and inference pipeline. We use the same batch size (200 or 400, accordingly), and adopt Adam optimizer with a learning rate of 1e-4 to optimize all methods. Each model is trained for 50 epochs, with early stopping after 10 epochs, and the best-performing checkpoint is used for testing. We report the average results across 3 runs. We also denote cases where the model failed due to either Out-Of-Time, where the model was unable to complete a run within 24 hours or out-of-memory on a 24GB GPU.

\subsection{Main Results}

\begin{figure*}[t]
    \centering
    \includegraphics[width=0.85\linewidth]{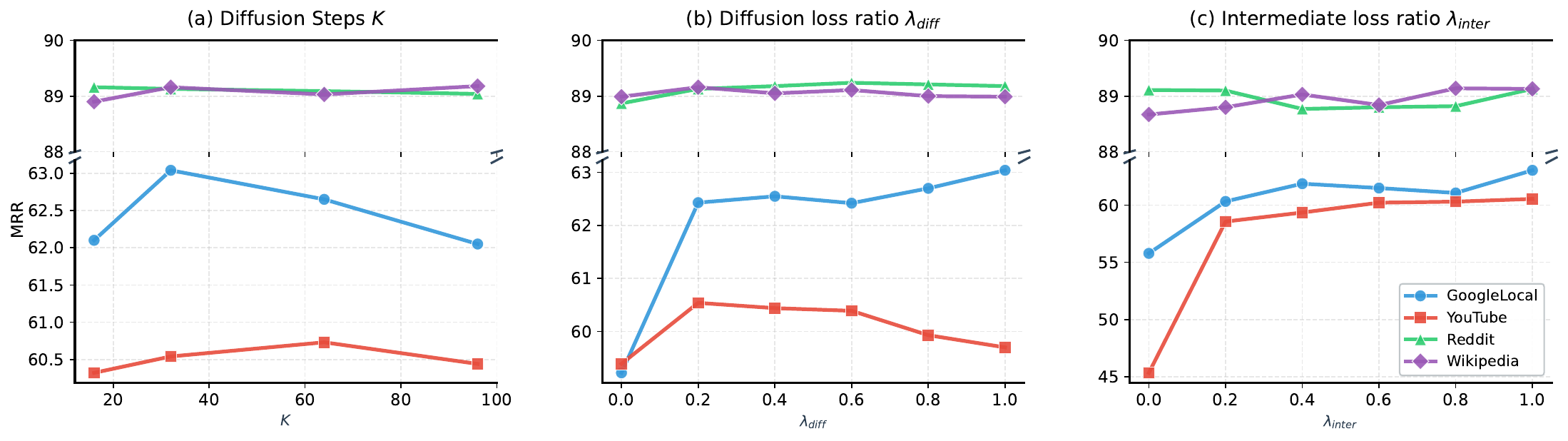}
    \includegraphics[width=0.85\linewidth]{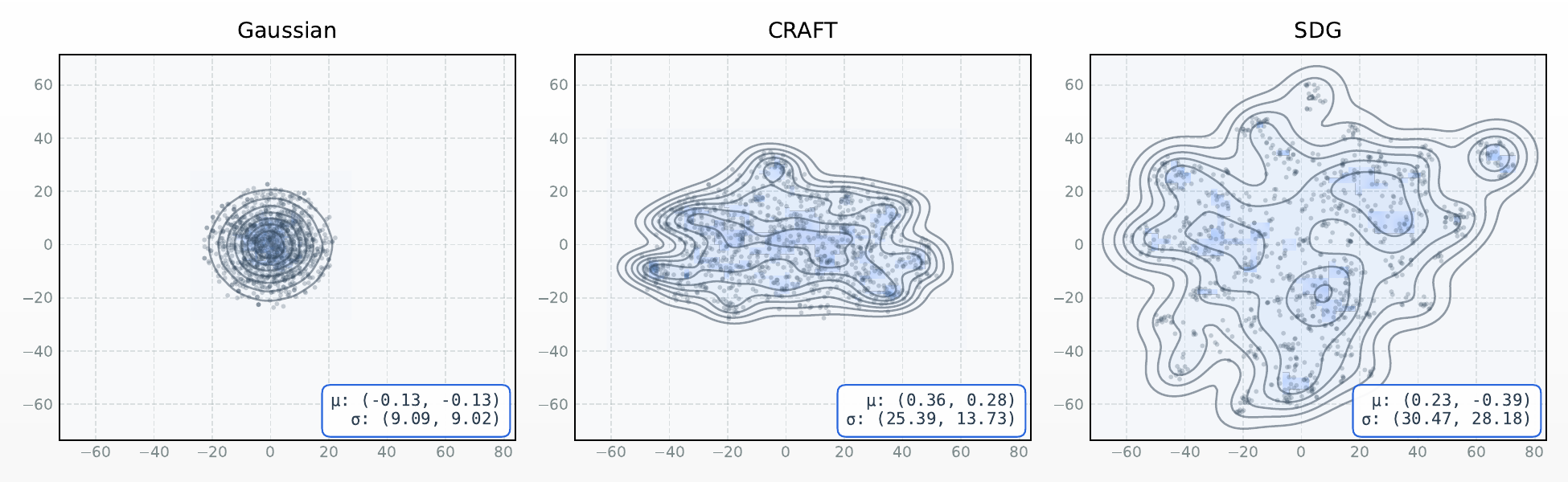}

    \caption{
    \textbf{(Top)} Performance of SDG with different hyperparameter settings, including (a) Diffusion Steps, (b) Diffusion Loss Ratio, and (c) Intermediate Loss Ratio.
    \textbf{(Bottom)} T-SNE visualization of learned node embeddings on GoogleLocal.
    }
    \label{fig:analysis}
\end{figure*}

Table \ref{tbl:seen} compares SDG with seven temporal graph baselines across five small-scale datasets, with Rel. Imprv and Abs. Imprv, indicating the relative and absolute improvements over the second-best baseline. Overall, SDG achieves the best results and second-best results compared to competing baselines. The most pronounced gains are observed on MOOC, where SDG improves the second-best model by 2.99\% in MRR and 1.59\% in HR@10. SDG slightly underperforms CRAFT and DyGFormer on Lastfm and UCI in HR@10 with 1.10\% and 3.27\% degradation. The reason may be attributed to the characteristics of these datasets, as they contain large volumes of repeated interactions, which reduces the benefit of modeling temporal distribution.

The performance of SDG and baseline methods on the TGB-Seq datasets is summarized in \autoref{tbl:unseen}. As these datasets contain non-repeating edges, many existing methods struggle in this setting. Across all five datasets, SDG consistently achieves the best performance, with CRAFT emerging as the strongest competing baseline. Overall, SDG improves MRR by 0.84\%–14.48\% and HR@10 by 1.59\%–8.72\%, with absolute gains of up to 7.92 MRR and 6.30 HR@10, most notably on GoogleLocal. On the two largest datasets (ML-20M and Taobao), DyGFormer and TGAT fail to scale due to prohibitive computational costs, whereas SDG maintains strong performance with high efficiency. These results demonstrate that SDG effectively captures long-range temporal dependencies across both recurring and non-recurring interaction regimes while remaining scalable.

\subsection{Ablation Study}

To evaluate the effectiveness of each design choice in SDG, we perform ablation studies in Table \ref{tab:ablation} with four variants: \textbf{(i)}. [w/o Seq] adds noise and denoise only the final destination embedding. \textbf{(ii)}. [w/o Diff] removes the diffusion module, only using the encoding module to make a prediction. \textbf{(iii)}.[MSE] replaces the proposed Cosine Loss with the standard MSE Loss. \textbf{(iv)}. [MLP] replaces the Cross Attention Transformer with an MLP as the denoising model.

From Table \ref{tab:ablation}, we observe that removing any component of SDG consistently degrades performance, confirming that each design choice contributes meaningfully to the model’s effectiveness. In particular, removing the sequence denoising module (w/o Seq) leads to clear drops in both metrics—for example, on GoogleLocal, MRR decreases from 62.60 to 55.27, highlighting the importance of modeling sequential distributions. Disabling the diffusion mechanism (w/o Diff) also causes notable performance degradation, demonstrating the role of diffusion in capturing uncertainty and complex interaction dynamics. Replacing the diffusion objective with an MSE loss further degrades performance by a substantial 12.32\% on GoogleLocal. Substituting the denoising network with a simple MLP causes substantial drops on 2 TGB-Seq datasets, with a slight gain on Wikipedia, suggesting that an MLP is only sufficient when temporal patterns are relatively simple, and most dependencies are already captured by the encoder.
\begin{table}[h]
\caption{Results (\%) of ablation experiments conducted for SDG on four datasets.}
\vspace{2mm}
\label{tab:ablation}
\centering
\resizebox{\linewidth}{!}{
\begin{tabular}{@{}ccccccc@{}}
\toprule
\textbf{Dataset} & \textbf{Metric} & \textbf{SDG} & \textbf{w/o Seq} & \textbf{w/o Diff} & \textbf{MSE} & \textbf{MLP} \\ \midrule
\multirow{2}{*}{\textbf{Wikipedia}} & MRR & \underline{89.16} & 88.63 & 89.11 & 89.05 & \textbf{89.40} \\
& HR@10 & \underline{91.45} & 90.92 & 91.41 & 91.32 & \textbf{91.69} \\ \midrule
\multirow{2}{*}{\textbf{Reddit}} & MRR & \textbf{89.24} & 89.12 & 89.13 & \underline{89.19} & 88.95 \\
& HR@10 & \textbf{94.86} & \underline{94.83} & 94.76 & 94.64 & 94.53 \\ \midrule
\multirow{2}{*}{\textbf{GoogleLocal}} & MRR & \textbf{62.60} & 55.27 & \underline{55.67} & 54.89 & 44.66  \\
& HR@10 & \textbf{78.54}  & \underline{73.61}  & 70.25 & 69.25  & 61.89 \\ \midrule
\multirow{2}{*}{\textbf{YouTube}} & MRR & \textbf{60.25} & 56.95 & \underline{58.59} & 58.31 & 51.62 \\
& HR@10 & \textbf{70.86} & 67.53 & 69.29 & \underline{69.53} & 61.74 \\
\bottomrule
\end{tabular}%
}
\end{table}

\subsection{Hyperparameter Sensitivity}

We first investigate three key parameters in our proposed methods: \textbf{(i)}. the total number of diffusion step ($K$), \textbf{(ii)}. the diffusion loss ratio $\lambda_{\text{diff}}$, and \textbf{(iii)}. the intermediate loss ratio $\lambda_{\text{inter}}$. The experiments are conducted in Figure \ref{fig:analysis} (Top).

As shown in Figure \ref{fig:analysis} (a), increasing the diffusion steps initially improves performance, while further increases may introduce redundant computations without providing additional modeling benefits. This suggests that choosing a moderate diffusion step, such as $K=32$, offers a good trade-off between performance and computational efficiency.
Figure \ref{fig:analysis} (b) demonstrates that introducing a non-zero diffusion loss weight consistently enhances performance, confirming the effectiveness of the diffusion objective. As
$\lambda_{\text{diff}}$ increases, MRR improves steadily; however, excessively large values may cause the reconstruction objective to dominate, thereby interfering with the ranking objective.
Finally, Figure \ref{fig:analysis} (c) shows that incorporating intermediate-position supervision significantly improves results, particularly on GoogleLocal and YouTube. An appropriate choice of $\lambda_{\text{inter}}$ helps the model learn more consistent temporal transition distributions along the destination sequence.

\subsection{Scalability and Running time}
\begin{figure}[h]
    \centering
    \includegraphics[width=1.0\linewidth]{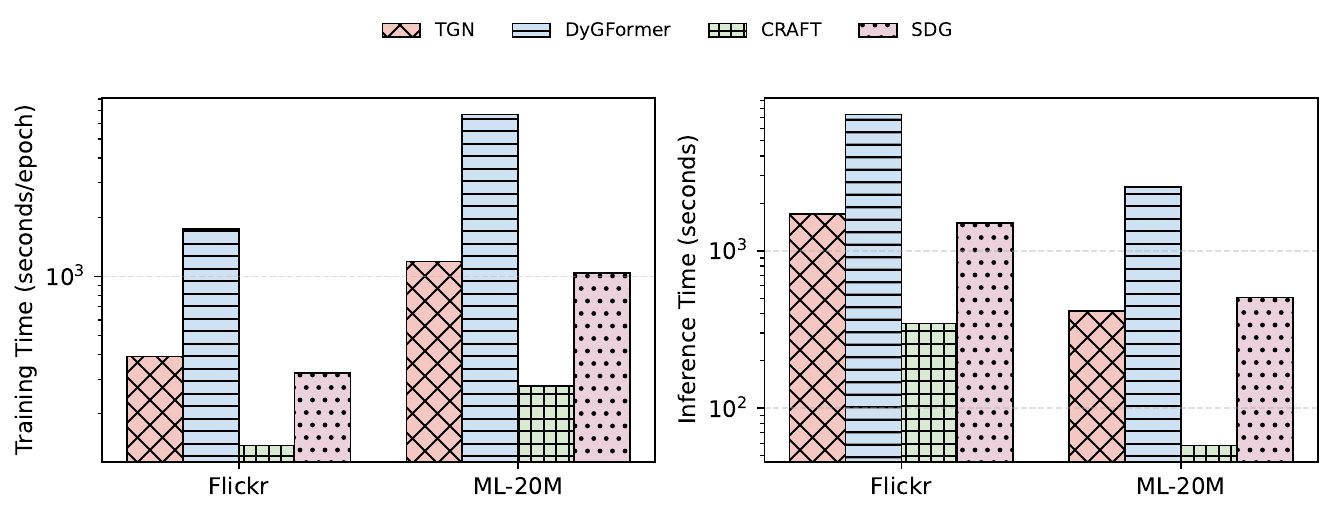}
    \caption{Training and Inference time comparison on two large-scale datasets Flickr and ML-20M.}
    \label{fig:scalability}
\end{figure}

We evaluate the efficiency of SDG by measuring both inference and
training times on two representative datasets of increasing scale: Flickr (7M edges), and ML-20M (18M edges). Figure \ref{fig:scalability} presents the training time in seconds per epoch and inference time in seconds for the test set.

Compared to traditional discriminative baselines, SDG performs multiple reverse diffusion steps during inference, introducing additional computations. However, with a moderate number of diffusion steps, SDG can maintain competitive inference time while delivering significantly stronger predictive performance. During training, SDG is the second fastest method per epoch. Overall, SDG remains more efficient than transformer-based methods such as DyGFormer, whose inference cost scales poorly with dataset size, demonstrating a favorable trade-off between efficiency and modeling capacity for large-scale temporal graphs.

\subsection{Noisy Scenario Evaluation}

We conduct a robustness test by randomly inserting 10\% to 60\% noisy edges with chronological timestamps during the evaluation.
\begin{figure}[h]
    \centering
    \includegraphics[width=1.0\linewidth]{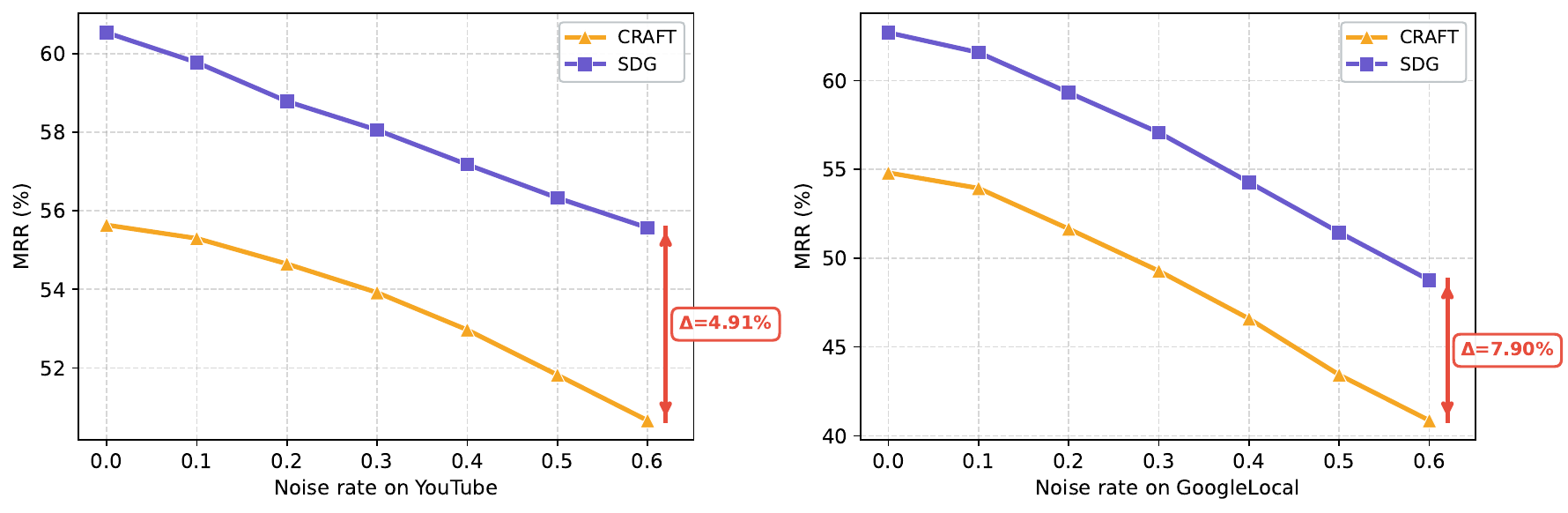}
    \caption{Inserting noisy edges from 10\% to 60\%.}
    \label{fig:robustness}
\end{figure}
In Figure \ref{fig:robustness}, SDG can exhibit competitive performance when the proportion of noisy edges increases. Compared to CRAFT, under the noisiest settings, SDG outperform by $4.9\%$ and $7.9\%$ on YouTube and GoogleLocal, respectively. This indicates that SDG can effectively handle noisy and uncertain interaction patterns.
\subsection{Embedding Visualization}
We provide a detailed visualization of the learned embedding on GoogleLocal of CRAFT and SDG in Figure \ref{fig:analysis} (Bottom). Due to the large number of nodes, we randomly select 10000 nodes as example. From the figure, we observe that CRAFT yields a more dispersed but still relatively smooth embedding manifold, though the distribution remains fairly homogeneous. In contrast, SDG produces the most diverse and structured embedding space, characterized by broader clusters and coverage. This richer geometry indicates that SDG captures more complex, heterogeneous interaction dynamics and uncertainty, aligning with its superior predictive performance on challenging temporal link prediction tasks.
\section{Conclusion}

In this work, we introduce Sequence Diffusion for Dynamic Graphs (SDG), a novel framework that bridges temporal graph learning with diffusion-based generative modeling. By casting temporal representation learning as a denoising process over historical neighbor interactions, SDG effectively captures fine-grained temporal patterns, long-range dependencies, and the inherent uncertainty of dynamic interactions. Through extensive experiments on diverse benchmark datasets, we demonstrated that SDG consistently outperforms state-of-the-art temporal graph models in future link prediction, while maintaining high efficiency for large-scale applications. For future work, we plan to extend SDG using different diffusion frameworks and explore its application to inductive settings. We believe our work highlights diffusion as a powerful new paradigm for modeling temporal link prediction in continuous-time dynamic graphs.

\section{Impact Statement}
This paper presents work whose goal is to advance the field of Machine Learning. There are many potential societal consequences of our work, none which we feel must be specifically highlighted here.


\bibliography{example_paper}
\bibliographystyle{icml2026}

\newpage
\appendix
\onecolumn

\section*{Appendix}
\section{Experiments Details}

\subsection{Dataset Details}
\label{sec:data_detail}
In our experiments, we adopt a diverse set of 10 datasets: five from TGB-Seq, and five commonly used datasets in the field. Table \ref{tbl:ds} presents the key statistics of these datasets for reference. Detailed descriptions of the datasets:

\textbf{Wikipedia\footnote{\url{https://zenodo.org/records/7213796\#.Y1cO6y8r30o}}.} \cite{jodie} A bipartite temporal graph of user–page edit interactions on Wikipedia, where nodes represent users and pages, and edges denote timestamped edits. Each interaction is associated with a 172-dimensional LIWC feature, and dynamic labels indicate whether a user is temporarily banned.

\textbf{Reddit\footnotemark[2].} \cite{jodie} A bipartite interaction dataset capturing users posting in subreddits over one month. Nodes correspond to users and subreddits, and edges represent timestamped posting events with 172-dimensional LIWC features and dynamic user ban labels.

\textbf{MOOC\footnotemark[2].} \cite{jodie} A bipartite network of online education activities, consisting of students and course content units (e.g., videos and assignments). Edges indicate student access events and are annotated with 4-dimensional features.

\textbf{LastFM\footnotemark[2].} \cite{jodie} A bipartite user–song interaction dataset recording music listening behavior over one month, where edges represent timestamped listening events.

\textbf{UCI\footnotemark[2].} \cite{uci} A temporal, non-attributed communication network among students at the University of California, Irvine, with interactions timestamped at second-level granularity.

\textbf{ML-20M.\footnote{\url{https://drive.google.com/drive/folders/1qoGtASTbYCO-bSWAzSqbSY2YgHr9hUhK}}.} A standard recommendation benchmark from MovieLens containing timestamped user–movie ratings. Following common practice, ratings are converted to implicit feedback, yielding a bipartite user–item graph for temporal interaction prediction.

\textbf{Taobao\footnotemark[3].} \cite{taobao_tgb} An e-commerce user behavior dataset containing timestamped click interactions between users and products over a nine-day period, represented as a bipartite temporal graph.

\textbf{GoogleLocal\footnotemark[3].} \cite{gl_tgb} A business review dataset derived from Google Maps, where user–business interactions are treated as implicit feedback. The resulting bipartite graph captures timestamped review events.

\textbf{YouTube\footnotemark[3].} \cite{yt_tgb} A non-bipartite “Who-To-Follow” social network derived from YouTube, where nodes are users and edges represent timestamped subscription relationships.

\textbf{Flickr\footnotemark[3].} \cite{flickr_tgb} A non-bipartite social network from Flickr, capturing timestamped friendship formations between users. Similar to YouTube, the task is to predict future user–user connections.

\begin{table}[h]
  \centering
  \caption{Datasets statistics: five TGB-Seq datasets, and five commonly used datasets. This table is partially adopted from TGB-Seq.}
  \label{tbl:ds}
  \resizebox{\linewidth}{!}{
  \begin{tabular}{lccccccc}
  \toprule
    \textbf{Dataset} & \textbf{Nodes (users/items)} & \textbf{Edges}  & \textbf{Timestamps} & \textbf{Repeat ratio(\%)}  &  \textbf{Bipartite} & \textbf{Domain}\\
    \midrule
    {ML-20M} & 100,785/9,646 & 14,494,325 & 9,993,250 & 0 & $\checkmark$ & Movie rating\\
    {Taobao} & 760,617/863,016 & 18,853,792 & 139,171 & 16.58 & $\checkmark$ & E-commerce interaction\\
    {GoogleLocal} & 206,244/267,336 & 1,913,967 & 1,771,060 & 0 &  $\checkmark$ & Business review \\
    {Flickr} & 233,836 & 7,223,559& 134 & 0 & $\times$ & Who-To-Follow \\
    {YouTube} & 402,422 & 3,288,028& 203 & 0 & $\times$ & Who-To-Follow \\
    \midrule
    {Wikipedia} & 8,227/1,000 & 157,474 & 152,757 & 88.41 & $\checkmark$ & Interaction \\
    {Reddit} & 10,000/984 & 672,447 & 669,065 & 88.32 & $\checkmark$ & Reply network \\
    {MOOC} & 7,047/97 & 411,749& 345,600 & 56.66 & $\checkmark$ & Interaction \\
    {LastFM} & 980/1,000 & 1,293,103 & 1,283,614 & 88.01 & $\checkmark$ & Interaction \\
    {UCI} & 1,899 & 59,835 & 58,911 & 66.06 & $\times$ & Social contact \\
  \bottomrule
  \end{tabular}
  }
\end{table}

\subsection{Baselines}
\label{sec:baseline_detail}
We briefly summarize the baseline methods evaluated in our experiments.

\textbf{JODIE} \cite{jodie} integrates a memory module and an aggregation module. The memory module encodes historical interactions into a compact latent state using an RNN, while the aggregation module projects this memory state into a time-aware node embedding via a temporal context vector.

\textbf{DyRep} \cite{dyrep} also adopts an RNN-based memory update mechanism but differs by applying temporal graph attention to transform interaction histories before updating memory states. Node embeddings are obtained directly from the memory states through an identity aggregation.

\textbf{TGN} \cite{tgn} employs a GRU to update node memories and uses temporal graph attention to aggregate historical neighbor information—including memory states, node features, edge attributes, and timestamps—into node representations.

\textbf{TGAT} \cite{tgat} introduces temporal graph attention without relying on a memory module, using stacked attention layers to aggregate time-aware neighbor information.

\textbf{GraphMixer} \cite{graphmixer} is a sequence-based approach that aggregates historical interactions using an MLP-Mixer architecture, without maintaining explicit memory states. It adopts a fixed time encoding scheme, which has been shown to outperform learnable temporal encodings in this setting.

\textbf{DyGFormer} \cite{dygformer} is a sequence-based method that employs a Transformer encoder for neighbor aggregation. It captures structural relationships between source and destination nodes by modeling their neighbor co-occurrence frequencies, effectively encoding shared interaction patterns through common neighbors.

\textbf{CRAFT} \cite{craft} introduces a cross-attention mechanism that enables target-aware modeling between candidate destinations and the source’s interaction history, addressing the challenge of unseen edge prediction.

\subsection{Implementation Details}
\label{sec:exp_detail}
\textbf{Experiment Environment} We conduct experiments on a server with AMD Ryzen 9 7950X with 16 cores and a NVIDIA RTX 4090 (24GB) GPU. The code is written in Python 3.11, and we use PyTorch 2.0.1 on CUDA 11.8 to train the model.

\textbf{Configuration Details.} For all baselines except CRAFT, we follow the official configurations of DyGLib\footnote{\url{https://github.com/yule-BUAA/DyGLib}} \cite{dygformer}, and TGB-Seq\footnote{\url{https://github.com/TGB-Seq/TGB-Seq}} \cite{tgb-seq}. For CRAFT \cite{craft}, we follow the original paper and disable shuffle-based training to ensure consistency with other temporal baselines that require the link stream to be processed in chronological order. This constraint leads to a noticeable performance drop compared to the original CRAFT results, but ensures a fair temporal evaluation. For SDG, we set the embedding size to 64 for small datasets and 128 for large ones, while the number of diffusion steps is searched over [32, 64, 96]. The loss weights $\lambda_{\text{diff}}$ and $\lambda_{\text{inter}}$ are tuned in the range [0.2, 0.4, $\dots$, 1.0]. We also tune the number of neighbors $L$ in the range of [30, 60, 90] for each source node. All hyperparameters are chosen based on validation performance, and the optimal configurations for each dataset are reported in Table \ref{tab:sdg_hyperparams}.

\begin{table}[h]
\centering
\caption{Optimal hyperparameters and embedding size $d$ for SDG, and the fixed batch size $B$ on various datasets.}
\vspace{2mm}
\label{tab:sdg_hyperparams}
\begin{tabular}{lccccccccc}
\toprule
Dataset 
& $L$
& $K$
& $\lambda_{\text{inter}}$
& $\lambda_{\text{diff}}$
& $N_{\text{layers}}$
& $B$
& $d$
& $\mathcal{L}_{task}$\\
\midrule
uci          & 30 & 32 & 1.0 & 0.2 & 1 & 200 & 64 & BCE  \\
wikipedia    & 30 & 96 & 1.0 & 0.2 & 1 & 200 & 64 & BCE  \\
reddit       & 30 & 32 & 1.0 & 0.6 & 1 & 200 & 64 & BCE  \\
mooc         & 30 & 32 & 1.0 & 1.0 & 2 & 200 & 64 & BCE  \\
lastfm       & 60 & 32 & 1.0 & 1.0 & 1 & 200 & 64 & BCE  \\
GoogleLocal  & 60 & 32 & 1.0 & 0.8 & 2 & 200 & 128 & BCE \\
Flickr       & 90 & 32 & 1.0 & 1.0 & 1 & 400 & 128 & BPR \\
YouTube      & 60 & 64 & 1.0 & 0.2 & 2 & 400 & 128 & BCE \\
Taobao       & 60 & 32 & 1.0 & 1.0 & 2 & 400 & 128 & BPR \\
ML-20M       & 60 & 32 & 1.0 & 1.0 & 2 & 400 & 128 & BCE \\
\bottomrule
\end{tabular}
\end{table}

\textbf{Loss Function.} We also test SDG Bayesian Personalized Ranking loss, which is widely used and well-suited for ranking tasks. We choose BPR as the main loss for Flickr and Taobao as it give better results compare to BCE Loss. The $\mathcal{L}_{task}$ for BPR version can be expressed as follows:
\begin{align}
\mathcal{L_{\text{task}}}
= &\underbrace{
    - \log \sigma(\hat{y}^+_{t, L} - \hat{y}^-_{t, L})
}_{\mathcal{L}_{\text{last}}}
+\lambda_{\text{inter}}\,
  \underbrace{
   \frac{1}{L - 1}\sum_{i=1}^{L-1}
      - \log \sigma(\hat{y}^+_{t, i} - \hat{y}^-_{t, i})
  }_{\mathcal{L}_{\text{inter}}},
\end{align}
where $\hat{y}^+_{t, L}$ and $\hat{y}^-_{t, i}$ denote the predicted scores for the positive and negative samples, respectively.

\textbf{Negative Sampling.} \cite{edgebank} proposed two challenging strategies for future link prediction: historical and inductive negative sampling. In this work, we adopt multiple negative samples per positive interaction and follow the random negative sampling protocol used in prior benchmarks. Empirically, using multiple random negatives per instance is sufficiently challenging and provides a fair and consistent basis for comparison across models.

\textbf{Settings for Noisy Evaluation.} As illustrated in Figure \ref{fig:robustness}, we assess model robustness by examining resistance to noise in both edges and timestamps. During training, models are learned under a transductive setting using random negative sampling. For evaluation, after neighbor sampling, we inject noise by randomly selecting $\sigma \times L$ positions within each sequence, where both the node identities and timestamps at these positions are randomly perturbed. The noise ratio $\sigma$ is varied from 0.1 to 0.6 to control the severity of corruption.

\textbf{Implementation for SDG under recurring datasets.} Based on Table \ref{tbl:ds}, we categorize datasets into recurring ones with repeat ratio $\ge 80\%$ and unseen ones with repeat ratio $\le 20\%$. For recurring datasets, we incorporate repeat-time encoding \cite{craft} into the scoring function in Equation \ref{eq:score}. For datasets with few repeated edges, we apply the default SDG model described in the main text, without repeat-time encoding.

\subsection{Experiment with point-wise metrics}

In the main text, we adopt MRR as the primary evaluation metric, as it is well-suited for ranking-based tasks with multiple negative samples and directly reflects the relative ordering of positive edges among negatives. In this section, we additionally report classification-oriented metrics, namely AP and ROC-AUC, following the random negative sampling protocol used in prior work \cite{dygformer}. As shown in Table \ref{tab:ap_auc}, SDG consistently outperforms competing methods across datasets, achieving the best overall average rank of 1.50. In particular, SDG ranks first on Reddit and MOOC, and second on Wikipedia and LastFM, demonstrating its strong and robust performance under diverse evaluation criteria.

\begin{table}[h]
\centering
\caption{Performance comparison across datasets using AP and ROC-AUC (\%).}
\label{tab:ap_auc}
\resizebox{\linewidth}{!}{
\begin{tabular}{llcccccccc}
\toprule
\textbf{Dataset} & \textbf{Metric} & \textbf{SDG} & \textbf{CRAFT} & \textbf{DyGFormer} & \textbf{GraphMixer} & \textbf{TGN} & \textbf{DyRep} & \textbf{JODIE} & \textbf{TGAT} \\
\midrule
\multirow{2}{*}{Wikipedia}
 & AP &
 \underline{98.73$\pm$0.03} & 97.61$\pm$0.14 & \textbf{99.08$\pm$0.07} & 97.25$\pm$0.03 & 98.45$\pm$0.06 & 94.86$\pm$0.06 & 96.50$\pm$0.14 & 96.94$\pm$0.06 \\
 & ROC-AUC 
 &  \underline{98.23$\pm$0.08} & 96.94$\pm$0.17 & \textbf{98.91$\pm$0.02} & 96.92 $\pm$0.03 &  98.37$\pm$0.07 & 94.37$\pm$0.09 & 96.33$\pm$0.07 & 96.67$\pm$0.07 \\
\midrule
\multirow{2}{*}{Reddit}
 & AP 
 & \textbf{99.46$\pm$0.01} & 99.21$\pm$0.01 & \underline{99.27$\pm$0.05} & 97.31$\pm$0.01 & 98.63$\pm$0.06 & 98.22$\pm$0.04 & 98.31$\pm$0.14 & 98.52$\pm$0.02 \\
 & ROC-AUC 
 & \textbf{99.38$\pm$0.02} & 99.03$\pm$0.01 &  \underline{99.15$\pm$0.01} & 97.17$\pm$0.02 & 98.60$\pm$0.06 & 98.17$\pm$0.05 & 98.31$\pm$0.05 &  98.47$\pm$0.02\\
\midrule
\multirow{2}{*}{LastFM}
 & AP 
 & \underline{93.10$\pm$0.01} & \textbf{93.32$\pm$0.01} & 92.99$\pm$0.20 & 75.99$\pm$0.22 & 78.24$\pm$2.39 & 70.43$\pm$3.42 & 69.23$\pm$0.75 & 73.00$\pm$0.35 \\
 & ROC-AUC 
 & \underline{92.23$\pm$0.11} & \textbf{92.51$\pm$0.02} & 92.05$\pm$0.10 & 73.53$\pm$0.12 & 78.47$\pm$2.94 &  71.16$\pm$1.89 & 70.49$\pm$1.66 & 71.59$\pm$0.18 \\
\midrule
\multirow{2}{*}{MOOC}
 & AP 
 & \textbf{96.05$\pm$0.12} & \underline{95.04$\pm$0.20} & 89.36$\pm$0.41 & 83.51$\pm$0.17 & 91.00$\pm$3.75 & 81.30$\pm$1.87 & 82.18$\pm$1.37 & 86.34$\pm$0.49 \\
 & ROC-AUC 
 & \textbf{96.18$\pm$0.08} & \underline{95.54$\pm$0.45} & 89.75$\pm$0.58 & 84.01$\pm$0.17 & 91.21$\pm$1.15 & 85.03$\pm$0.58 & 83.81$\pm$2.09 & 87.11$\pm$0.19 \\
 \midrule
 \multicolumn{2}{c}{Avg. Rank} & \textbf{1.50} & 2.50 & 2.50 & 6.00 & 4.00 & 7.75 & 7.00 & 5.25 \\
\bottomrule
\end{tabular}
}
\end{table}

\subsection{More Ablation Study}
We conduct more experiments related to the noise schedule and the neighbor sequence length of the SDG framework.

\begin{figure}[h]
    \centering
    \includegraphics[width=0.8\linewidth]{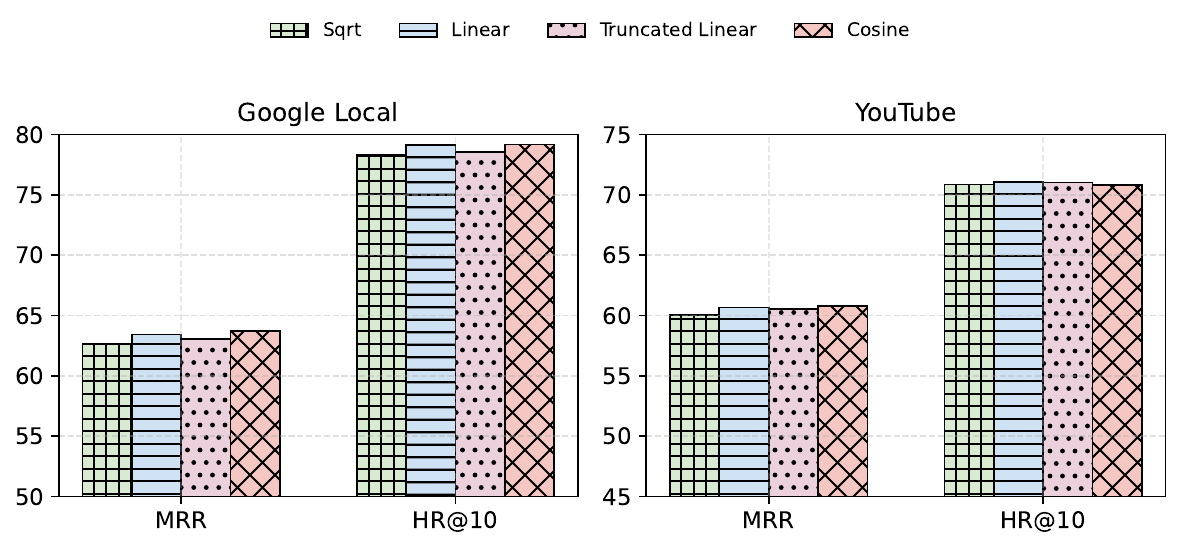}
    \caption{Performance of SDG with different noise schedules on GoogleLocal and YouTube.}
    \label{fig:noise_schedule}
\end{figure}

\textbf{Noise Schedule} Figure \ref{fig:noise_schedule} investigates the influence of noise scheduler $\alpha(s)$on SDG by comparing four commonly used schedules: linear, cosine, square-root (sqrt), and truncated linear on GoogleLocal and YouTube.
Across both datasets, the cosine schedule consistently achieves the strongest performance in terms of both MRR and HR@10, while sqrt performs the worst.
This can be attributed to its smoother noise decay for the cosine schedule, which allows for more stable transitions during the reverse diffusion process and facilitates more accurate recovery of the clean destination sequence.
While different noise schedules lead to noticeable performance differences, the overall variation remains relatively small, suggesting that SDG is not overly sensitive to the specific choice of noise schedule.

\textbf{Neighbor Length} Figure 6 illustrates the impact of the neighbor sequence length $L$ on SDG’s performance. On both YouTube and GoogleLocal, increasing $L$ from 30 to 60 consistently improves MRR and HR@10, indicating that leveraging a longer interaction history helps the model capture richer temporal dependencies and user behavior patterns.
However, when $L$ is further increased to 90, performance slightly degrades on both datasets. This suggests that excessively long histories may introduce outdated or noisy
interactions, which can dilute the relevance of recent behavior and hinder prediction accuracy.
Overall, $L=60$ achieves the best balance between incorporating sufficient historical context and avoiding noise from less informative past interactions.

\begin{figure}[h]
    \centering
    \includegraphics[width=0.8\linewidth]{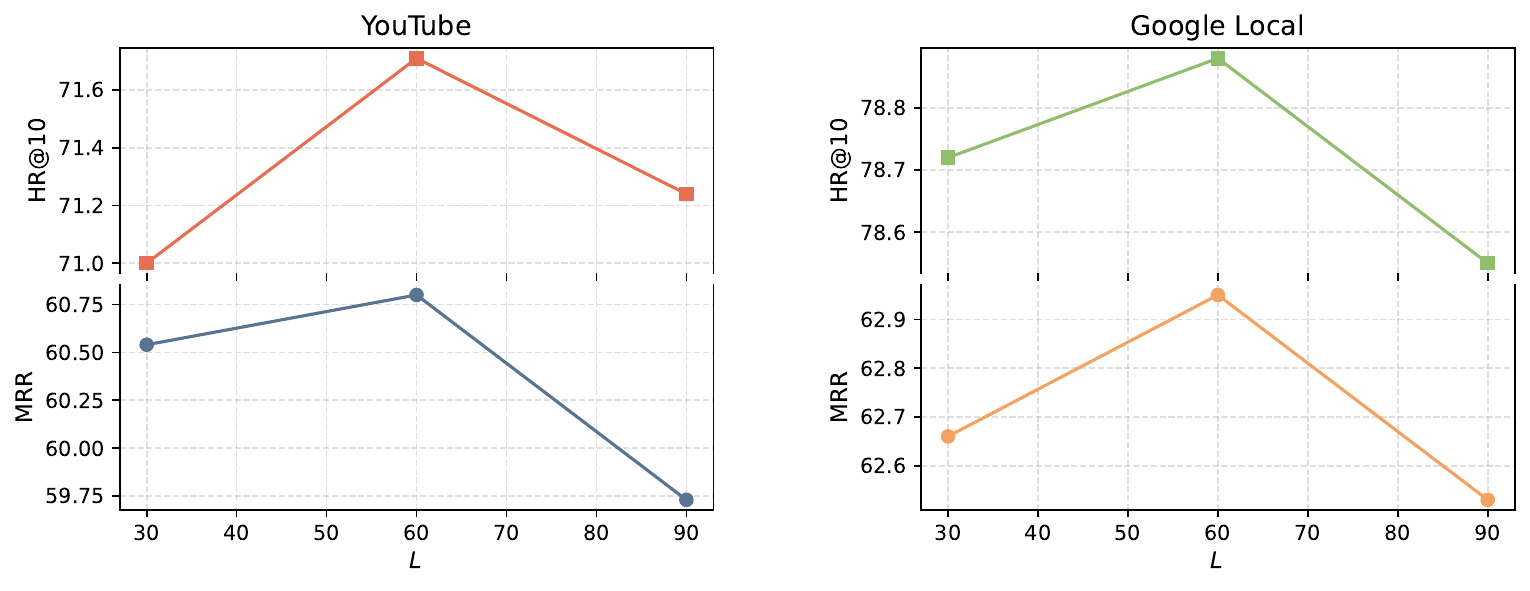}
    \caption{Performance of SDG with different neighbor length $L$}
    \label{fig:nl}
\end{figure}

\textbf{Memory Usage} Table 7 compares GPU memory consumption across methods on five datasets of increasing scale. CRAFT consistently exhibits the lowest memory footprint, reflecting its lightweight architecture. SDG requires moderately more memory than CRAFT due to the additional denoising model, but it remains substantially more memory-efficient than baseline models such as DyGFormer and GraphMixer. Notably, on large-scale datasets like ML-20M and Taobao, SDG reduces memory usage by more than 2–4× compared to DyGFormer and TGN, demonstrating that the sequence diffusion design achieves a favorable balance between modeling capacity and memory efficiency for large temporal graphs.
\begin{table}[h]
\centering
\caption{Comparison of GPU memory consumption (MiB).}
\label{tab:gpu_memory}

\begin{tabular}{lccccc}
\toprule
\textbf{Datasets} & \textbf{TGN} & \textbf{GraphMixer} & \textbf{DyGFormer} & \textbf{CRAFT} & \textbf{SDG} \\
\midrule
GoogleLocal & 4336 & 3960 & 3688 & 1356 & 1972   \\
YouTube & 4866 & 4804 & 5182 & 1908 & 2828  \\
Flickr & 6606 & 7274 & 7652 & 1372 & 3404   \\
ML-20M & 16492  & 11966 & 16002 & 602 & 3110  \\
Taobao & 16064 & 11672 & 16254 & 3178 & 4858 \\
\bottomrule
\end{tabular}
\end{table}

\section{Time Complexity Analysis}

We analyze the time complexity of performing temporal link predictions for a batch of source nodes and their corresponding set of candidate destinations. Let $B$ denote the batch size, $L$ the number of historical neighbors considered per node, $M$ the number of candidate destinations per source, $d$ the embedding or feature dimensions, and $N_{deg}$ the average node degree on the temporal graph. The total time complexity generally consists of two parts: 1) extracting historical neighbors and 2) computing edge likelihoods. Table \ref{tab:complexity} summarizes the time complexity of different TGNN methods.

\begin{table}[h]
\centering
\caption{Big-$\mathcal{O}$ time complexity of different temporal graph learning methods, where only the dominant term is retained for clarity.}
\label{tab:complexity}
\resizebox{0.8\linewidth}{!}{
\begin{tabular}{lcc}
\toprule
\textbf{Methods} & \textbf{Extract historical neighbors} & \textbf{Compute edge likelihoods} \\
\midrule
TGAT~\cite{tgat} 
& $B(1+M)(1+L)\log N_{deg}$ 
& $B(1+M)L^{2}d^{2}$ \\

TGN~\cite{tgn} 
& $B(1+M)\log N_{deg}$ 
& $B(1+M)Ld^{2}$ \\

GraphMixer~\cite{graphmixer} 
& $B(1+M)\log N_{deg}$ 
& $B(1+M)Ld^{2}$ \\

DyGFormer~\cite{dygformer} 
& $B(1+M)\log N_{\text{deg}}$ 
& $B(1+M)(Ld^{2}+L^{2}d)$ \\

CRAFT~\cite{craft} 
& $B\log N_{\text{deg}}$ 
& $BMd^{2} + BLd^{2} + BMLd$ \\

SDG 
& $B\log N_{\text{avg}}$ 
& $KB(L^2d+Ld^2) + BMd$ \\
\bottomrule
\end{tabular}
}
\end{table}

For the extraction of historical neighbors, we consider the common strategy of retrieving the $L$ most recent neighbors, following the widely adopted implementation in DyGLib. The dominant cost arises from locating the most recent neighbor among all existing neighbors via binary search, taking $O(\text{log}N_{deg})$ per query. Since existing methods typically gather neighbors for both source and destination nodes, this step takes $O(B(1 + M) \text{log}N_{deg})$. CRAFT and SDG only considers the neighbors of the source, reducing the complexity to $O(B \text{log} N_{deg})$. This reduction is particularly beneficial in real-world scenarios, where it is common to evaluate many candidate destinations per source, making efficient neighbor retrieval critical for scalable inference.

For predicting temporal link, TGN, TGAT, GraphMixer, and DyGFormer compute node representations by aggregating each node’s historical neighbors, followed by an MLP for prediction. The time complexity heavily depends on the method's architecture. For example, DyGFormer employs Transformer encoder on both the destination set and the source node, making the complexity $O(B(1+M)(Ld^{2}+L^{2}d))$. For SDG, we take into account the total diffusion time step during the inference process, resulting in the complexity of $(KB(L^2d + Ld^2))$ with $K$ being the diffusion step. Moreover, the dot product also costs an additional $O(BMd)$ to rank the candidates. 


\section{ELBO Derivation for Sequence Diffusion}
\label{sec:elbo}


Let $p_{\text{data}}(\mathbf{X}_{j}^0, \mathbf{Z})$ denote the empirical distribution of position $j$ in the clean sequences $\mathbf{X}_j^0$ and the encoded historical context $\mathbf{Z} = \mathbf{Z}_{1:L}(S_{u,t})$. Our ideal goal is to learn a conditional generative model $p_\theta(\mathbf{X}_{1:L}^0 \mid \mathbf{Z})$ that matches the true conditional distribution
$p_{\text{data}}(\mathbf{X}_{1:L}^0 \mid \mathbf{Z})$ by minimizing the
negative log-likelihood (NLL):
\begin{equation}
\mathcal{L}_{\text{ideal}} = \sum_{j=1}^L
\mathbb{E}_{\mathbf{Z} \sim p_{\text{data}}} \Big[H\big(p_{\text{data}}(\mathbf{X}_j^0 \mid \mathbf{Z}), p_\theta(\mathbf{X}_j^0 \mid \mathbf{Z})\big)\Big] = \sum_{j=1}^L
\mathbb{E}_{(\mathbf{X}_j^0,\mathbf{Z}) \sim p_{\text{data}}}
\big[- \log p_\theta(\mathbf{X}_j^0 \mid \mathbf{Z})\big].
\label{eq:ideal-nll}
\end{equation}
Following \cite{ddpm}, we can use the Evidence Lower Bound (ELBO) method
\begin{equation}
\begin{aligned}
\mathbb{E}_{(\mathbf{X}^0,\mathbf{Z}) \sim p_{\text{data}}}
\big[
- \log p_\theta(\mathbf{X}^0 \mid \mathbf{Z})
\big] &\le
\mathbb{E}_{p_{\text{data}}(\mathbf{X}^0,\mathbf{Z})}
\Bigg(
\mathbb{E}_{q(\mathbf{X}^{1:K} \mid \mathbf{X}^0)}
\Big[
\log
\frac{
q(\mathbf{X}^{1:K} \mid \mathbf{X}^0)
}{
p_\theta(\mathbf{X}^{0:K} \mid \mathbf{Z})
}
\Big]
\Bigg) \\
&=
\mathbb{E}_{q(\mathbf{X}^{0:K}, \mathbf{Z})}
\Big[
\log
\frac{
q(\mathbf{X}^{1:K} \mid \mathbf{X}^0)
}{
p_\theta(\mathbf{X}^{0:K} \mid \mathbf{Z})
}
\Big]
\;:=\;
\mathcal{L}_{\text{diff}} .
\end{aligned}
\end{equation}

We can expand and use the Bayes formula

\begin{equation}
\begin{aligned}
\mathcal{L}_{\text{diff}}
&=
\mathbb{E}_{q(\mathbf{X}^{0:K}, \mathbf{Z})}
\Bigg[
\log
\frac{
q(\mathbf{X}^{1:K} \mid \mathbf{X}^0)
}{
p_\theta(\mathbf{X}^{0:K} \mid \mathbf{Z})
}
\Bigg] \\
&=
\mathbb{E}_{q(\mathbf{X}^{0:K}, \mathbf{Z})}
\Bigg[
\log
\frac{
\prod_{k=1}^{K} q(\mathbf{X}^k \mid \mathbf{X}^{k-1})
}{
p(\mathbf{X}^K)
\prod_{k=1}^{K} p_\theta(\mathbf{X}^{k-1} \mid \mathbf{X}^k, \mathbf{Z})
}
\Bigg] \\
&=
\mathbb{E}_{q(\mathbf{X}^{0:K}, \mathbf{Z})}
\Bigg[
\log \frac{q(\mathbf{X}^K \mid \mathbf{X}^0)}{p(\mathbf{X}^K)}
+
\sum_{k=2}^{K}
\log
\frac{
q(\mathbf{X}^{k-1} \mid \mathbf{X}^k, \mathbf{X}^0)
}{
p_\theta(\mathbf{X}^{k-1} \mid \mathbf{X}^k, \mathbf{Z})
}
-
\log p_\theta(\mathbf{X}^0 \mid \mathbf{X}^1, \mathbf{Z})
\Bigg].
\end{aligned}
\end{equation}

The first term of the ELBO is a constant under a fixed prior, and the third term corresponds to the reconstruction loss. Focusing on the second term, we expand the expectation and identify it as a sum of KL divergences between the true reverse posterior and the learned reverse process at each diffusion step.

\begin{equation}
\begin{aligned}
\mathbb{E}_{q(\mathbf{X}^{0:K}, \mathbf{Z})}
\Bigg[
\sum_{k=2}^{K}
\log
\frac{
q(\mathbf{X}^{k-1} \mid \mathbf{X}^k, \mathbf{X}^0)
}{
p_\theta(\mathbf{X}^{k-1} \mid \mathbf{X}^k, \mathbf{Z})
}
\Bigg]
&=
\int q(\mathbf{X}^{0:K}, \mathbf{Z})
\sum_{k=2}^{K}
\log
\frac{
q(\mathbf{X}^{k-1} \mid \mathbf{X}^k, \mathbf{X}^0)
}{
p_\theta(\mathbf{X}^{k-1} \mid \mathbf{X}^k, \mathbf{Z})
}
\, d\mathbf{X}^{0:K} \\
&=
\sum_{k=2}^{K}
\mathbb{E}_{q(\mathbf{X}^k, \mathbf{X}^0, \mathbf{Z})}
\Big[
D_{\mathrm{KL}}\big(
q(\mathbf{X}^{k-1} \mid \mathbf{X}^k, \mathbf{X}^0)
\;\|\;
p_\theta(\mathbf{X}^{k-1} \mid \mathbf{X}^k, \mathbf{Z})
\big)
\Big].
\end{aligned}
\end{equation}
Assuming Gaussian parameterizations with fixed variance, each KL term reduces (up to constants) to a squared error between the posterior mean and the model prediction. Since the posterior mean is linear in the clean sample, the objective can be reparameterized as a reconstruction loss on $x^0$, as:
\begin{equation}
\begin{aligned}
\mathcal{L}_{\text{diff}}
&=
\sum_{k=2}^{K}
\mathbb{E}_{q(\mathbf{X}^k, \mathbf{X}^0, \mathbf{Z})}
\Bigg[
\frac{1}{2\sigma_k^2}
\big\|
\tilde{\boldsymbol{\mu}}_k(\mathbf{X}^k, \mathbf{X}^0)
-
\boldsymbol{\mu}_\theta(\mathbf{X}^k, k, \mathbf{Z})
\big\|^2
\Bigg] \propto
\mathbb{E}_{(k,\, \mathbf{X}^0,\, \boldsymbol{\epsilon},\, \mathbf{Z})}
\big[
\|\mathbf{X}^0 - \hat{\mathbf{X}}^0(\mathbf{X}^k, k, \mathbf{Z})\|^2
\big].
\end{aligned}
\end{equation}
Adding the cumulative sum of each position, we get:
\begin{equation}
\begin{aligned}
\mathcal{L}_{\text{diff}}
&= \sum_{j=1}^L \mathbb{E}_{(k,\, \mathbf{X}_{j}^0,\, \boldsymbol{\epsilon},\, \mathbf{Z})}
\big[
\|\mathbf{X}_{j}^0 - \hat{\mathbf{X}}^0_{j}(\mathbf{X}_{j}^k, k, \mathbf{Z})\|^2
\big].
\end{aligned}
\end{equation}

\textbf{Equivalence of MSE and Cosine Error under Unit-Norm Constraint.} To mitigate sensitivity to embedding norms and dimensionality, we instead consider a cosine-based reconstruction error.
The cosine similarity between the predicted clean embedding
$\hat{\mathbf{X}}^0$ and the ground-truth embedding $\mathbf{X}^0$ is defined as:
\begin{equation}
\mathcal{S}(\hat{\mathbf{X}}^0, \mathbf{X}^0)
=
1 - \cos(\hat{\mathbf{X}}^0, \mathbf{X}^0).
\label{eq:cosine_error}
\end{equation}
\begin{equation}
\cos(\hat{\mathbf{X}}^0, \mathbf{X}^0)
=
\frac{
\langle \hat{\mathbf{X}}^0, \mathbf{X}^0 \rangle
}{
\|\hat{\mathbf{X}}^0\|_2 \, \|\mathbf{X}^0\|_2
}.
\label{eq:cosine_sim}
\end{equation}
When both $\hat{\mathbf{X}}^0$ and $\mathbf{X}^0$ are normalized to have unit
$\ell_2$ norm, the squared Euclidean distance between them can be written as:
\begin{equation}
\begin{aligned}
\|\hat{\mathbf{X}}^0 - \mathbf{X}^0\|_2^2
&=
(\hat{\mathbf{X}}^0 - \mathbf{X}^0)^\top
(\hat{\mathbf{X}}^0 - \mathbf{X}^0) \\
&=
\|\hat{\mathbf{X}}^0\|_2^2 + \|\mathbf{X}^0\|_2^2
- 2 \langle \hat{\mathbf{X}}^0, \mathbf{X}^0 \rangle \\
&=
2 \big(1 - \cos(\hat{\mathbf{X}}^0, \mathbf{X}^0)\big).
\end{aligned}
\label{eq:mse_cosine_equiv}
\end{equation}
Thus, under the unit-norm constraint, minimizing the MSE reconstruction term in the ELBO is equivalent to minimizing the cosine error, up to a constant scaling factor of 2. This equivalence justifies replacing the Euclidean reconstruction loss with a cosine-based objective, which is invariant to embedding scale and better aligned with ranking-based temporal link prediction. Accordingly, we adopt the following cosine reconstruction loss for diffusion training:
\begin{equation}
\mathcal{L}_{\text{diff}}
=
\frac{1}{L}
\sum_{i=1}^{L}
\bigl(
1 - \cos(\hat{\mathbf{X}}^0_i, \mathbf{X}^0_i)
\bigr)^2.
\label{eq:final_cosine_loss}
\end{equation}

\section{More Analysis of Embedding}
We reduce the dimensionality of the learnable node embeddings using T-SNE \cite{tsne} to visualize the underlying distribution of the embedding space learned by SDG. Due to the
large number of nodes in ML-20M and Taobao, we randomly select 10000 nodes as example. Then, we apply Gaussian kernel density estimation \cite{kde} to analyze the density distribution of reduced embeddings and visualize the results using contour plot.

\begin{figure*}[h]
    \centering
    \includegraphics[width=0.8\linewidth]{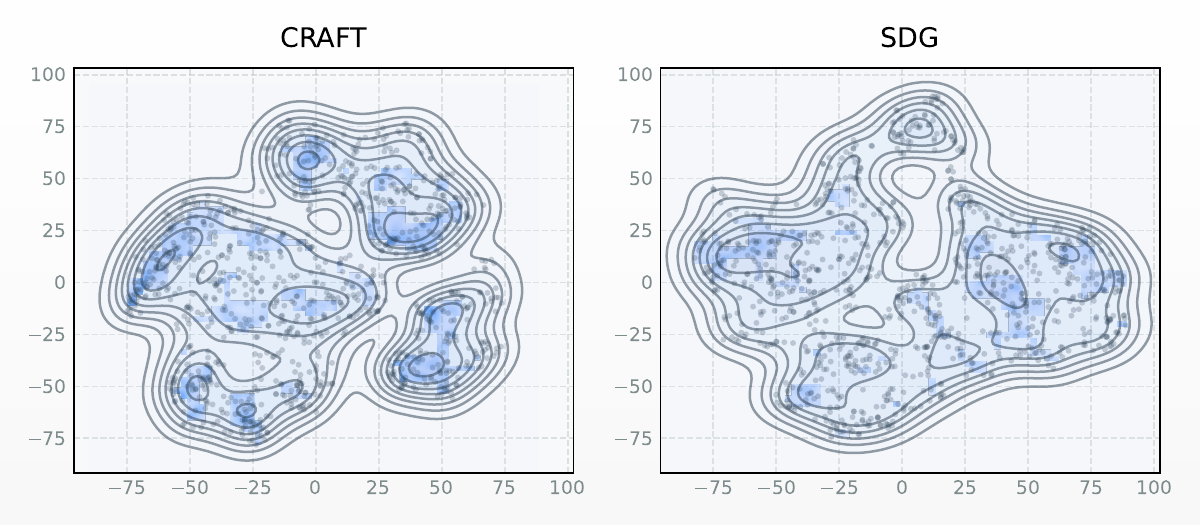}
    \includegraphics[width=0.8\linewidth]{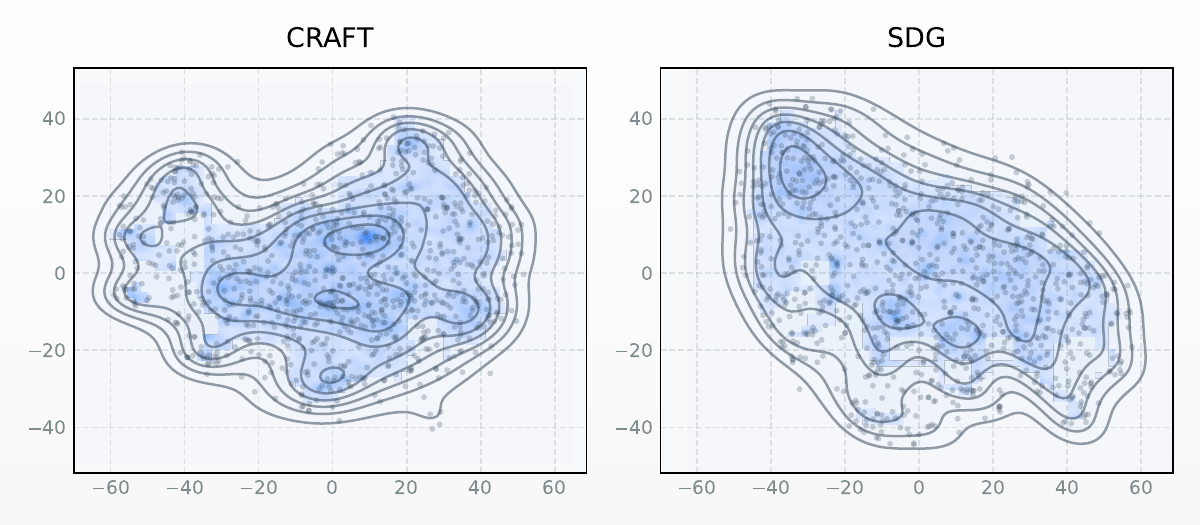}

    \caption{
     T-SNE visualization of learned node embeddings on ML-20M (Top) and Taobao (Bottom).
    }
    \label{fig:more_analysis}
\end{figure*}

From Figure \ref{fig:more_analysis}, we observe that SDG produces node embeddings with more coherent and smoother global geometry across both datasets. On ML-20M, SDG yields better-separated yet internally compact clusters. On Taobao, SDG embeddings exhibit a more unified and anisotropic structure, suggesting reduced noise and higher consistency in learned representations. Overall, SDG captures the underlying data distribution more faithfully, leading to better ranking performance.

\section{Limitations and Future Work}

\textbf{Additional Features.} The current implementation of SDG does not utilize explicit node or edge features. This design choice is motivated by the observation that most temporal graph benchmarks either lack such features or offer features that provide limited additional benefit for future link prediction, as reported in prior work \cite{graphmixer, craft}. Nevertheless, additional feature information can be utilized to guide the generation in the conditional diffusion framework. In particular, node features can be concatenated with learnable node embeddings, while edge features can be integrated into neighbor sequence representations or combined with positional encoding to provide edge-specific information for context guidance.

\textbf{Transductive and Inductive Future Link Prediction.}
In this paper, we do not explicitly separate the transductive and inductive tasks like previous works. All datasets follow the original training, validation, and test splits provided by existing benchmarks or are split chronologically. As a result, our evaluation naturally encompasses both settings.
SDG is primarily designed for transductive future link prediction, where sufficient interaction history is available to model node behavior. In inductive scenarios with little or no historical data, prediction becomes inherently difficult due to the cold-start problem, which typically requires additional node features or auxiliary information. We therefore treat inductive link prediction as a separate challenge beyond the scope of this work.

\textbf{Limitations.} While SDG achieves strong empirical performance, it also introduces extra computational overhead compared to purely discriminative temporal models. Even with a small number of diffusion steps, sequence-level denoising, Transformer-based conditioning, and auto-regressive supervision increase both training and inference cost. Moreover, SDG is specifically designed for future link prediction and does not directly support other tasks such as dynamic node classification. Our framework focuses on generating the destination node embeddings, without maintaining dynamic node representations over time.

\textbf{Future Works.} Several directions remain open for future exploration. First, extending SDG to explicitly address inductive/cold-start future link prediction is an important avenue, particularly by incorporating node attributes. Second, improving computational efficiency is a promising direction, for example, by reducing the number of diffusion steps, using modern techniques such as distillation, one-step generation, etc. In addition, extending sequence diffusion beyond future link prediction to support other dynamic graph tasks, such as node classification or representation tracking over time, would further broaden the applicability of the framework. Finally, exploring other diffusion model-based frameworks for temporal link prediction, such as Discrete-based Diffusion or Diffusion Bridge.

\section{Algorithm}
Algorithms \ref{alg:inference} and \ref{alg:training} formalize the inference and training procedures for the proposed Sequence Diffusion for Dynamic Graphs (SDG) model. In the inference procedure (Algorithm \ref{alg:inference}), SDG first encodes the historical interaction sequence $S_{u,t}$ to obtain contextual representations. The target sequence is then initialized with Gaussian noise. Next, the reverse diffusion process learned from data is applied repeatedly to obtain a denoised destination sequence. Finally, the prediction scores are computed based on the reconstructed sequence.
In the training procedure (Algorithm \ref{alg:training}), SDG builds historical and target interaction sequences, adds Gaussian noise to the target sequence at a randomly chosen diffusion step, and trains the denoising network to predict the clean sequence given the historical context.
The final objective function integrates the ranking loss and the diffusion reconstruction loss, allowing for end-to-end optimization of context encoding and sequence generation.
\begin{algorithm}[h]
\caption{Inference of Sequence Diffusion}
\label{alg:inference}
\begin{algorithmic}[2]
\REQUIRE Source node $u$, historical sequence $S_{u, t}$, prediction time $t$, candidate set $\mathcal{C}_{u, t}$, diffusion steps $K$, parameters $\theta$
\STATE $\mathbf{X}^K_{1:L} \sim \mathcal{N}(0, I)$.
\STATE $\mathbf{Z}_{1:L}(S_{u, t}) = \text{Transformer}(\mathbf{H}_{1:L}(S_{u, t}); \mathbf{M})$.
\FOR{$k = K, K-1, \dots, 1$}
    \STATE $\hat{\mathbf{X}}^0_{1:L} = f_\theta(\mathbf{Z}_{1:L}(\mathbf{S}_{u, t}), \mathbf{X}_{1:L}^{k}, k)$.
    \STATE $\hat{\mathbf{X}}^{k-1}_{1:L} = \frac{\sqrt{1 - \beta^k}(1 - \bar{\alpha}^k)}{1 - \bar{\alpha}^k} \mathbf{X}^k_{1:L} + \frac{\alpha^{k - 1}\beta^k}{1 - \bar{\alpha}^k} \hat{X}^0_{1:L} + \sigma_k \epsilon$.
    \STATE $\mathbf{X}^{k-1}_{1:L} = \hat{\mathbf{X}}^{k-1}_{1:L}$.
\ENDFOR
\STATE Compute scores $\hat{\mathbf{y}}_t$ from Equation \eqref{eq:score} \end{algorithmic} \end{algorithm}

\begin{algorithm}[h]
\caption{Training of Sequence Diffusion}
\label{alg:training}
\begin{algorithmic}[1]
\REQUIRE Batch of interactions $\{(u, v, t)\}$, parameters $\theta$
\FOR{each interaction $(u, v, t)$ in the batch}
    \STATE Construct $S_{u,t}$ and $\mathcal{T}_{u,t}$
    \STATE $\mathbf{Z}_{1:L}(S_{u, t}) = \text{Transformer}(\mathbf{H}_{1:L}(S_{u, t}); \mathbf{M})$.
    \STATE $\mathbf{X}^0 = \mathbf{H}(\mathcal{T}_{u,t})$.
    \STATE Sample $\boldsymbol{\epsilon} \sim \mathcal{N}(0, I)$.
    \FOR{$i = 1, 2, \dots, L$}
        \STATE $\mathbf{X}_{i}^k = \sqrt{\bar{\alpha}^{k}}\,\mathbf{X}^0_i + \sqrt{1 - \bar{\alpha}^{k}} \boldsymbol{\epsilon}_i$.
    \ENDFOR
    \STATE $\hat{\mathbf{X}}^0_{1:L} = f_\theta(\mathbf{Z}_{1:L}(\mathbf{S}_{u, t}), \mathbf{X}_{1:L}^{k}, k)$
    \STATE Compute scores $\hat{\mathbf{y}}_t$ from Equation \eqref{eq:score}
    \STATE Compute $\mathcal{L} = \mathcal{L}_{\text{task}} + \lambda_{\text{diff}}\mathcal{L}_{\text{diff}}$.
\ENDFOR
\STATE Update $\theta$ with gradient descent on $\mathcal{L}$.
\end{algorithmic}
\end{algorithm}

\end{document}